%% file: main.tex
\documentclass[11pt]{article}

\usepackage{geometry}
\usepackage[]{natbib}

\usepackage{graphicx}

\usepackage{hyperref}

\usepackage{amssymb}
\usepackage{amsmath,bm,mathrsfs,amsfonts,algorithm}

\usepackage[noend]{algpseudocode}

\usepackage{listings}
\makeatletter
\gdef\lst@numberfirstlinefalse{\global\let\lst@ifnumberfirstline\iffalse}
\lst@AddToHook{Init}{\global\let\lst@ifnumberfirstline\iftrue}
\makeatother

\usepackage{soul}

\usepackage{tikz}
\usetikzlibrary{positioning,fit}
\usetikzlibrary{decorations.text}
\usetikzlibrary{arrows}
\usetikzlibrary{shapes}

\input{img/settings.tex}

\newtheorem{example}{Example}

\usepackage{xcolor}

\usepackage{color}

\definecolor{mygreen}{rgb}{0,0.6,0}
\definecolor{mygray}{rgb}{0.5,0.5,0.5}
\definecolor{kwrds}{rgb}{0.99,0.0,0.0}
\definecolor{backcolour}{rgb}{0.95,0.95,0.92}

\newcommand{\weight}[1]{\textcolor{mygreen}{\mathbf{#1}}}
\newcommand{\scalar}[1]{\textcolor{mygreen}{#1}}
\newcommand{\wc}[2]{\textcolor{#1}{\mathbf{#2}}}
\newcommand{\conj}{\textcolor{kwrds}{\scalebox{1.2}{, }}}
\newcommand{\dw}{\textcolor{red}{:}}

\newcommand{\impl}{\textcolor{blue}{ {\scalebox{1.5}{{:-}}} }}

\newenvironment{centeredornot}
    {\begin{center}
    \begin{tabular}{c}
    }
    {
    \end{tabular} 
    \end{center}
    }

\input{img/settings.tex}
\begin{document}



\title{Beyond Graph Neural Networks\\with Lifted Relational Neural Networks
}


\author{
Gustav Sourek
\footnote{
corresponding author: {souregus@fel.cvut.cz}           
}
\and
        Filip Zelezny      \and
        Ondrej Kuzelka 
}

\date{%
    Department of Computer Science\\
    Faculty of Electrical Engineering\\
    Czech Technical University in Prague\\
}



\maketitle

\begin{abstract}

We demonstrate a declarative differentiable programming framework based on the language of Lifted Relational Neural Networks, where small parameterized logic programs are used to encode relational learning scenarios. When presented with relational data, such as various forms of graphs, the program interpreter dynamically unfolds differentiable computational graphs to be used for the program parameter optimization by standard means. Following from the used declarative Datalog abstraction, this results into compact and elegant learning programs, in contrast with the existing procedural approaches operating directly on the computational graph level.
We illustrate how this idea can be used for an efficient encoding of a diverse range of existing advanced neural architectures, with a particular focus on Graph Neural Networks (GNNs). Additionally, we show how the contemporary GNN models can be easily extended towards higher relational expressiveness. In the experiments, we demonstrate correctness and computation efficiency through comparison against specialized GNN deep learning frameworks, while shedding some light on the learning performance of existing GNN models.

\end{abstract}

\section{Introduction}
\label{sec:intro}

It has been recently proposed {by several authors} that incorporating logic reasoning capabilities into neural networks is crucial to achieve more powerful AI systems~\citep{marcus2020next,de2020statistical,lamb2020graph}. Indeed, we see a rising interest in enriching deep learning models with certain facets of symbolic AI, ranging from logical entailment~\citep{evans2018can}, rule learning~\citep{Evans2017}, and solving combinatorial problems~\citep{palm2018recurrent}, to proposing differentiable versions of a whole Turing machine~\citep{Graves2014,graves2016hybrid}.
However, similarly to the Turing-completeness of recurrent neural networks, the expressiveness of these advanced neural architectures is not easily translatable into actual learning performance, as their optimization tends to be often prohibitively difficult~\citep{lipton2015critical}.

There has been a long stream of research in neural-symbolic integration~\citep{Bader2005a,garcez2019neural}, traditionally focused on emulating logic reasoning within neural networks~\citep{KBANN,smolensky1990tensor,Botta,LiyaDing}. The efforts eventually evolved from propositional~\citep{KBANN,garcez1999connectionist} into full first order logic settings, mapping logic constructs and semantics into respective tensor spaces and optimization constraints~\citep{Serafini2016,dong2019neural,marra2020relational}. Traditionally, the neural-symbolic works focus on providing various perspectives into the correspondence between symbolic and sub-symbolic representations and computing, targeting again mostly novelty and theoretical expressiveness rather than practical learning applications.

From the bottom-up practical perspective, there has been a continuous effort of applying neural network learning to increasingly complex relational data~\citep{BlockeelB,dash2018large,Kazemi2018}. While learning from relational data has been traditionally dominated by approaches rooted in relational logic~\citep{muggleton1994inductive} and its probabilistic extensions~\citep{BLP,MLN,problog}, the neural networks offer highly efficient latent representation learning, which is beyond capabilities of the symbolic systems. Neural networks on the other hand have traditionally been based on fixed tensor representations, which cannot explicitly capture the unbounded, dynamic and irregular nature of the relational data.

Recursive~\citep{socher2013recursive} and Graph neural networks (GNNs)~\citep{scarselli2008graph} introduced a highly successful paradigm shift by moving from fixed neural architectures to dynamically constructed computation graphs, directly following the structural bias presented by the input examples. As opposed to the recurrent~\citep{Graves2014} and ``tensorization''~\citep{garcez2019neural} neural-symbolic approaches, this enables to exploit the structural properties of the data more efficiently, as they are simply directly coded into the very structure of the model, similarly to the original propositional neural-symbolic integration methods~\citep{KBANN}. Consequently, GNNS achieved remarkable success in a wide range of tasks~\citep{zhou2018graph}.

In targeting integration of deep and relational learning, one of the core desired properties for an integrated system is to keep expressiveness of both the worlds as a special case. While much focus has been devoted to keep the expressiveness of the logic reasoning, considerably less attention was put on the neural models, the expressiveness and modern variations of which are mostly ignored by the integrated systems~\citep{manhaeve2018deepproblog}. 

In this paper, we show how to use simple relational logic programs to capture advanced convolutional neural architectures in a tightly integrated and exact manner.
Particularly, we use the language of Lifted Relational Neural Networks (LRNNs)~\citep{sourek2018lifted}, and demonstrate that a wide range of neural models, ranging from simple perceptrons to complex GNNs, can be elegantly and efficiently covered -- not only from the perspective of expressiveness but, importantly, from the practical point of view
\footnote{Code to reproduce experiments from this paper is available at \url{https://github.com/GustikS/GNNwLRNNs}. The~LRNNs framework itself can then be found at \url{https://github.com/GustikS/NeuraLogic}.}.
We further show how to easily extend the basic GNN idea into some of the most contemporary GNN architectures, and \textit{beyond}.

The paper is structured as follows.
Firstly, we introduce the necessary preliminaries of logic and deep learning in Section~\ref{sec:back}. In Section~\ref{sec:lrnns}, we introduce the language of LRNNs, which we use throughout the paper. Subsequently, we illustrate LRNNs on a range of example models in Section~\ref{sec:examples}. Capturing and extending GNNs is then detailed in Section~\ref{sec:GNNs}. In Section~\ref{sec:experiments}, we demonstrate practicality and efficiency of the approach. We then discuss related works in Section~\ref{sec:related} and conclude in Section~\ref{sec:concl}.

\section{Background}
\label{sec:back}

Here we introduce the {necessary} preliminaries of (i) logic programming and (ii)~deep learning, which we seek to integrate.

\subsection{Logic}
\label{sec:back:logic}

\paragraph{Syntax:}
A \textit{relational} logic theory is a set of formulas formed from constants, variables, and predicates \citep{FOL}. Constant symbols represent objects in the domain of interest (e.g.\ $\textit{hydrogen}_1$) and will be written in lower-case. Variables  (e.g.\ $\textit{X}$) range over the objects in the domain and will be written with a capitalized first letter. Predicate symbols represent relations among objects in the domain or their attributes. A {\em term} may be a constant or variable. An {\em atom} is a predicate symbol applied to a tuple of terms (e.g.\ $\textit{bond}(X,\textit{hydrogen}_1)$). Formulas are constructed from atoms using logical connectives of $\vee$ and $\wedge$~\citep{FOL}. A {\em ground term} is a term containing no variables. A {\em ground atom}, also called \textit{proposition}, is an atom having only ground terms as arguments (e.g.\ $\textit{bond}(\textit{oxygen}_1,\textit{hydrogen}_1)$). A {\em literal} is an atom or a negation of an atom. A clause is a universally quantified disjunction of literals\footnote{Note we do not write the universal quantifiers explicitly in this paper.}. A clause with exactly one positive literal is a {\em definite clause}. A definite clause with no negative literals (i.e. consisting of just one literal) is called a {\em fact}. A definite clause $h \vee \neg b_1 \vee \dots \vee \neg b_k$ can also be written as an implication $h \leftarrow b_1 \wedge \dots \wedge b_k$. The literal $h$ is then called {\em head} and the conjunction $b_1 \wedge \dots \wedge b_k$ is called {\em body}. We will often call definite clauses, which are not facts, {\em rules}. A set of such rules is then commonly called a \emph{logic program}.

\paragraph{Semantics:}
The \emph{Herbrand base} of a set of relational formulas $\mathcal{P}$ is the set of all ground atoms which can be constructed using the constants and predicates that appear in this set. A {\em Herbrand interpretation} of $\mathcal{P}$, also called a {\em possible world}, is a mapping that assigns a truth value to each element from $\mathcal{P}$'s Herbrand base. We say that a possible world $I$ satisfies a ground atom $F$, written $I\models F$, if $F\in I$. The satisfaction relation is then generalized to arbitrary ground formulas in the usual way.
A set of ground formulas is satisfiable if there exists at least one possible world in which all formulas from the set are true; such a possible world is called a {\em Herbrand model}. Each set of definite clauses has a unique Herbrand model that is minimal w.r.t.\ the subset relation, called its \textit{least Herbrand model}. The least Herbrand model of a finite set of ground definite clauses can be constructed in a finite number of steps using the {\em immediate-consequence operator}~\citep{ico}. This immediate consequence operator is a mapping $T_p$ from Herbrand interpretations to Herbrand interpretations, defined for a set of ground definite clauses $\mathcal{P}$ as $T_p(I) = \{h \,|\, (h \leftarrow b_1 \wedge \dots \wedge b_k) \in \mathcal{P}, \{b_1,..., b_k\} \subseteq I\}$. In other words, the operator $T_p$ expands the current set of true atoms (i.e.\ the current Herbrand interpretation $I$) with their immediate consequences as prescribed by the rules in $\mathcal{P}$.

Now consider a set of non-ground definite clauses $\mathcal{P}$. The \emph{grounding} of a clause $\alpha$ from $\mathcal{P}$ is the set of ground clauses $G(\alpha) = \{\alpha\theta_1,...,\alpha\theta_n\}$ where $\theta_1,...,\theta_n$ is the set of all possible substitutions, each mapping the variables occurring in $\alpha$ to constants appearing in $\mathcal{P}$. Note that if $\alpha$ is already ground, its grounding is a singleton. The grounding of $\mathcal{P}$ is given by $G(\mathcal{P}) = \bigcup_{\alpha\in\mathcal{P}} G(\alpha)$. The least Herbrand model of $\mathcal{P}$ is then defined as the least Herbrand model of $G(\mathcal{P})$.
In practice, most of the rules in the grounding $G(\mathcal{P})$ will be irrelevant, as their body can never be satisfied. The \emph{restricted grounding} limits the grounding to those rules which are ``active'', i.e.\ whose body is satisfied in the least Herbrand model $\mathcal{H}$. It is defined by $G^R(\mathcal{P}) = \{h\theta \leftarrow b_1\theta \wedge \dots \wedge b_k\theta \,|\, (h \leftarrow b_1 \wedge \dots \wedge b_k) \in \mathcal{P}\mbox{ and } \{ h\theta, b_1\theta, \dots, b_k\theta \} \subseteq \mathcal{H} \}$.


\subsubsection{Logic Programming}
\label{sec:back:logic:prog}
Logic programming is a declarative programming paradigm for computation with logic programs. In this paradigm, definite clauses are used to express facts and rules about a domain, and the computation is then carried out by the means of logical inference. 
Syntactically, the rules in the program $h \leftarrow b_1 \wedge \dots \wedge b_k$ are commonly written as

\begin{centeredornot}
\begin{lstlisting}[mathescape=true]
h $\impl$ b$_1$ $\conj\dots\conj$ b$_k$.
\end{lstlisting}
\end{centeredornot}

\noindent where each ``$\conj$'' stands for conjunction, and ``{\small $\impl$}'' replaces the implication, which reads right-to-left. Facts are then simply rules with no body.

Particularly, we consider the language of Datalog~\citep{unman1989datalog}, a restricted function-free subset of Prolog~\citep{bratko2001prolog}. Datalog is a domain specific language used in advanced deductive database engines. In contrast with Prolog, Datalog is a truly declarative language, where the order of clauses does not influence execution, and it is also guaranteed to terminate. Separate efficient theorem proving engine can then be used for computing the execution inferences~\citep{bancilhon1985magic}.

Importantly, it is still a \textit{relational} language, and one can thus use variables in the clauses, enabling to compose general and reusable programming patterns, such as
\begin{center}
\begin{tabular}{c}
\begin{lstlisting}[mathescape=true]
h(X) $\impl$ edge(X,Y)$\conj$node(Y).
\end{lstlisting}
\end{tabular}
\end{center}
which will then be automatically bound to a multitude of ground data structures via different substitutions for the variables \{X,Y\}. We will further extensively use this non-ground expressiveness while extending Datalog towards differentiable programming\footnote{This is a distinguishing feature from many other \textit{procedural} differentiable programming languages, such as PyTorch or TensorFlow, which are effectively propositional in this sense.}.

\paragraph{Query:} Similarly to querying a standard database with SQL, one provides a \textit{query} atom ($q$) to execute a Datalog program towards inference of a specific target, e.g.
\begin{center}
\begin{tabular}{c}
\begin{lstlisting}[mathescape=true]
h($oxygen_1$)?
\end{lstlisting}
\end{tabular}
\end{center}
which then drives the theorem proving engine to find a model $I \models q$ of the given logic program. If successful, the result of the execution is then the query together with the possible world $I$ used for its derivation. Note that there may be multiple worlds that model $q$.



\subsection{Deep Learning}
\label{sec:back:neural}

Deep learning is a machine learning approach characterized by using multi-layered neural network models. A neural network is a parameterized computation graph, particularly a data-flow graph where the data flowing through the edges are being successively transformed by numeric operations represented by the nodes (neurons). A neural layer is a set of neurons residing at the same depth in the directed data-flow graph. A multi-layered (``deep'') neural network is a graph with more than two such layers. 

The data are commonly represented as (fixed-size) tensors, and the operations are commonly differentiable non-linear functions. Owing to the differentiability of the functions, the parameters of a graph, commonly associated with the edges, can be efficiently trained by gradient-descend routines.
Due to the increasingly complex nature of the computation graphs and the utilized operations, the field has been recently also referred to as \textit{differentiable programming}\footnote{Note however that, despite being theoretically Turing-complete (e.g. recurrent neural networks), the models themselves are rarely as expressive in practice as standard programming languages used for their creation.}.

\textit{Neural architectures} are common design patterns used in creation of the computation graphs for specific types of problems. Here we briefly review some of the most common and successful neural architectures used in deep learning~\footnote{We introduce these architectures explicitly, despite being commonly known, as we further work with them in detail as differentiable Datalog programs.}.

\subsubsection{Multi-layer Perceptrons}
\label{sec:back:mlp}

A multi-layered perceptron (MLP) is the original and most common neural architecture. It is a directed feed-forward data-flow graph. Moreover, the interconnections between nodes in subsequent layers commonly follow the ``fully-connected'' pattern (a complete bipartite graph). Consequently, assuming the common vector form of the input data, the computation graph can be efficiently reduced to a linear series of dense matrix multiplications, each followed by an element-wise application of a non-linear function, such as logistic sigmoid ($\sigma$) or rectified linear unit ($ReLU$). 

The main idea behind MLPs is ``representation learning'' of the input data, often referred to as \textit{embedding}, where one can think of outputs of the individual layers as transformed representations of the input, each extracting gradually more expressive information w.r.t. the output learning target.

\subsubsection{Convolutional Networks}
\label{sec:back:cnns}

A convolutional neural network (CNN) is also a feed-forward architecture, characterized by utilizing particular operations in one or more sub-parts of the computational graph. The specific operations are commonly referred to as ``\textit{convolution}'' (filtering) and ``\textit{pooling}''. Given a vector input of size $n$, the convolutional filter (kernel) will also be represented by a vector of size $k < n$, which is then successively element-wise multiplied with all the $k$-length subsequences of the input vector, to produce $n-k+1$ scalar values. The resulting values are commonly referred to as ``feature-maps''. The second operation is the pooling, which aggregates values from predefined spatial sub-regions of the input values (feature-maps) into a single output through application of some (non-parameterized) aggregation function such as mean ($avg$) or maximum ($max$). The layers of these operations can then be mixed together with the previously introduced layers from MLPs in various combinations.

The main idea behind the convolution operation is the application of the very same parameterized function over different regions of the input. This enables to \textit{abstract} away common patterns out of different sub-parts of the input representation. The main idea behind the pooling operation is to enforce \textit{invariance} w.r.t. translation of the inputs.


\subsubsection{Recursive and Recurrent Networks}
\label{sec:back:rnns}

A \textit{Recursive} Neural Network (RNN)\footnote{Note that the abbreviation is also used for the recurrent neural networks, in this paper however, we use it solely to refer to \textit{recursive} networks.} is a neural architecture which differs significantly from the previous in that the exact form of the computation graph is not given in advance. Instead, the computation graph structure directly follows the structure of an input example, which takes the form of a $k-$regular tree. This enables to learn neural networks directly from differently-structured regular tree examples, as opposed to the fixed-size tensors which can be seen as graphs with completely regular grid topologies.

The leaf nodes in the computation tree then represent the input data, each of which is associated with a feature vector (embedding). Every $k$ leaf nodes are consequently combined by a given operation to compute the representation for their common parent node. This combining operation then continues recursively for all interior nodes, until the representation for the root node is computed, which forms the output of the model. Similarly to the convolution in CNNs, the parameterized combining operation over the children nodes remains the same over the whole tree~\citep{socher2013reasoning}\footnote{In some works, this architecture is further extended to use a set of different parameterizations, depending for instance on given types associated with the nodes, such as types of constituents in constituency-based parse trees.}.

The main idea behind recursive networks is that neural learning can be extended towards structured data by generating a \textit{dynamic} computation graph for each individual example tree. The learning then exploits the convolution principle to discover the underlying \textit{compositionality} of the learning representations in recursive structures.

The basic form of a commonly known \textit{Recurrent} Neural Network~\citep{lipton2015critical} can then be seen as a ``restriction'' of the idea to sequential structures, i.e. linear chains of input nodes\footnote{We note that modern recurrent architectures use additional computation constructs to store the hidden state, such as the popular LSTM cells, which are more complex and do not directly follow from the input structure.}. The computation graph is then successively unfolded along the input sequence to compute the hidden representation for each node based on the previous node's representation and the current node features (current input). The main idea behind recurrent networks is that the hidden representation can store a sort of \textit{state} of the computation.

\subsection{Graph Neural Networks}
\label{sec:back:gnn}

Graph Neural Networks (GNN)\footnote{recently more popular in the form of ``Graph Convolutional Networks'', which slightly differ from the original GNN proposal~\citep{scarselli2008graph}, but share the general principles discussed.} can be seen as a further extension of the principle to completely irregular graph structures~\citep{bronstein2017geometric}. Similarly to the recursive networks, they dynamically unfold the computational graph from the input structure for the purpose. However, GNN is a multi-layered feed-forward neural architecture, where the structure of each layer $i$ exactly follows the structure of the \textit{whole} input graph. Every node $v$ in the graph can now be associated with a feature vector (embedding), forming the input layer $h(v)^{(0)} = features(v)$. Interestingly, however, this is not necessary in general, as the variance in the graph topologies of the individual examples can already provide enough discriminative information on its own. For computation of the next layer $i+1$ representations of the nodes, each node in the graph updates its own representation by {aggregating} representation vectors of the adjacent nodes (``message passing'') via some parameterized update operation. GNNs again exploit the convolution idea, while the same operation is again applied uniformly over the whole graph. Note that in contrast to recursive networks, a different parameterization is typically used at each layer.

The computation at each layer $i$ can be possibly divided into two steps~\citep{xu2018powerful}, where we firstly \textit{aggregate} the hidden representations $h(u)$ of the node's $v$ neighbors $u \in \mathcal{N}(v)$ to obtain some activation value as

$$act^{(i)}(v) = aggregate^{(i)}(\{h(u)^{(i-1)} : u \in \mathcal{N}(v) \})$$
and then we \textit{combine} this activation value with the node's $v$ own representation $h(v)$, to obtain its new updated representation to be used in the next layer as

$$h^{(i)}(v) = combine^{(i)} (h(v)^{(i-1)}, act^{(i)}(v))$$
This general principle covers a wide variety of the proposed GNN models, which then reduces to the choice of particular ``aggregate and combine'' operations. For instance in GraphSAGE~\citep{hamilton2017inductive}, the operations are

$$ act^{(i)}(v) = max \{ReLU(W \cdot h^{(i-1)}(u)) | u \in \mathcal{N}(v)\}$$ 
and 
$$h^{(i)}(v) = W_f \cdot [(h(v)^{(i-1)}, act^{(i)}(v)]$$
while in the popular Graph Convolutional Networks~\citep{kipf2016semi}, these can be even merged into a single step as

$$h^{(i)}(v) = ReLU(W \cdot avg\{  h^{(i-1)}(u) | u \in \mathcal{N}(v) \cup \{v\}\})$$
and the same generic principle applies to many other GNN works~\citep{xu2018representation,gilmer2017neural,xu2018powerful}.

GNNs can be directly utilized for both graph-level as well as node-level classification tasks.
For output prediction on the level of individual nodes, we simply apply some activation function on top of its last layer representation $query(v) = \sigma(h(v)^{(n)})$. For predictions on the level of the whole graph $\mathcal{G}$, all the node representations need to be aggregated by some pooling operation such as $query(\mathcal{G}) = \sigma(avg\{h^{(n)}(v) | v \in \mathcal{G}\})$.

By following the same pattern at each layer $i$, the computation will produce increasingly more aggregated representations, since at layer $i$ each node effectively aggregates representations from its ``$i$-hops'' neighborhood. Intuitively, the GNN inference can thus be seen as a continuous version of the popular Weisfeiler-Lehman algorithm~\citep{weisfeiler2006construction} for calculating graph fingerprints used for refutation checking in graph isomorphism testing.

A large number of different variants of the original GNNs~\citep{scarselli2008graph} have been proposed, recently achieving state-of-the-art empirical performance in many tasks~\citep{wu2019comprehensive,zhou2018graph}. In essence, each introduced GNN variant came up with a certain combination of common activation and aggregation functions, and/or proposed extending the architecture with additional connections~\citep{xu2018representation} or layers borrowed from other neural architectures~\citep{velivckovic2017graph,li2015gated}, nevertheless they all share the same introduced idea of successive aggregation of node representations. For a general overview, we refer to~\citep{wu2020comprehensive,zhou2018graph}.

\paragraph{Spectral GNNs:}

Here we discussed ``spatially'' represented graphs and operations. However, some GNN approaches represent the graphs and the convolution operation in spectral, Fourier-domain~\citep{wu2020comprehensive}. There the update operation is typically conveyed in the matrix form as
$$H^{(i)} = f( \hat{A} \times H^{(i-1)} \times W_{i-1})$$
where $\hat{A}$ is an altered\footnote{e.g. $\hat{A} = D^{-\frac{1}{2}} (A+I) D^{-\frac{1}{2}}$, where $D$ is the diagonal node-degree matrix and $I$ is an identity matrix, such as in the original Graph Convolutional Networks~\citep{kipf2016semi}.} adjacency matrix of the graph, encoding the respective neighborhoods, $H^{(i)}$ contains the successive hidden node representations at layer $i$, and $W_i$ are the learnable parameters at each layer. However we note that, not considering the specific normalizations and approximations used, these again follow the same ``aggregate and combine'' principles, and can be rewritten accordingly~\citep{xu2018powerful}. While theoretically substantiated in graph signal processing, spectral GNN models are generally inadvisable as they introduce substantial limitations in terms of efficiency, learning, generality, and flexibility~\citep{wu2020comprehensive}, and we do not consider them further in this paper.

\paragraph{Knowledge Base Embeddings:}

Knowledge Base Embeddings (KBEs) are a set of approaches designed for the task of knowledge base completion (KBC)~\citep{kadlec2017knowledge}, i.e. predicting existing (missing) edges in large knowledge graphs. Particularly, these methods approach the task through learning of a distributed representation (embedding) for the nodes. In multi-relational graphs, a representation of the edge (relation) can also be added, forming a commonly used triplet representation of $({o}bject,{r}elation,{s}ubject)$. To predict the probability of a given edge in the knowledge graph, KBEs then choose one of a plethora of functions designed to \textit{combine}\footnote{Note that there is no need for the ``aggregate'' operation in KBEs.} the three embeddings from the underlying triplet~\citep{kadlec2017knowledge}.

\section{The Language of Lifted Relational Neural Networks}
\label{sec:lrnns}

In this paper we follow up on the work of Lifted Relational Neural Networks (LRNNs)~\citep{sourek2015lifted} which have been introduced as a framework for \textit{templated} modeling of diverse neural architectures oriented to relational data. It can be understood as a differentiable version of simple Datalog programming, where the templates, encoding various neuro-relational architectures, take the form of parameterized logic programs. It differs from the commonly used frameworks, such as PyTorch or Tensorflow, in its declarative, relational nature, enabling one to abstract away from the procedural details of the underlying computational graphs even further. We explain principles of this abstraction in the following subsections.

\subsection{Syntax: Weighted Logic Programs}
\label{sec:syntax}
The syntax of LRNNs is derived directly from the Datalog~\citep{unman1989datalog} language (Section~\ref{sec:back:logic}), which we further extend with numerical parameters. Note that this has been exploited in many previous works, where the parameters can signify values associated with facts~\citep{bistarelli2008weighted} or rules~\citep{eisner2010dyna}. Such extensions are typically designed to integrate standard statistical (or probabilistic~\citep{de2007problog}) modelling techniques with the high expressiveness of relational representation and reasoning. In this work we seek to integrate Datalog with deep learning, for which we allow \textit{each literal} in each clause of the logic program to be associated with a tensor weight. A parameterized program, formed by a multitude of such weighted rules, then declaratively encodes all computations to be performed in a given learning scenario.
For clarity of correspondence with standard (neural) learning scenarios, we here further split\footnote{Note that this split is not necessary in general, and the template can also contain facts, as well as the learning examples may contain rules, such as in ILP learning scenarios.} the program into unit clauses (facts), constituting the learning examples, and definite clauses (rules), constituting the learning template.

\subsubsection{Learning Examples}


The learning examples contain factual description of a given world. For their representation we use weighted ground facts. A learning example is then a set ${E} = \{(V_1, e_1),\dots,(V_j,e_j)\}$, where each $V_i$ {is} a real-valued tensor and each $e_i$ is a ground fact, i.e. expression of the form
\begin{centeredornot}
\begin{lstlisting}[mathescape=true]
$\weight{V_1}$:: p$_1(c^1_1,\dots,c^1_q)$.
$\dots$
$\weight{V_j}$:: p$_n(c^n_1,\dots,c^n_r)$.
\end{lstlisting}
\end{centeredornot}
where $p_1,\dots,p_n$ are predicates with corresponding arities $q,\dots,r$, and $c_i^j$ are arbitrary constants.
{S}tandard logical representation {is} then a special case where each $\weight{V_i} = 1$\footnote{Since we consider a close world assumption (CWA) and least Herbrand model, one does not enumerate false facts with zero value.}. One can either write $\scalar{1}$\textcolor{red}{::}{$carbon(c_1$)} or ommit the weight and write $bond(c_1,o_2)$. The values do not have to be binary and can represent a ``degree of truth'' to which a certain fact holds, such as $\scalar{0.4}$\textcolor{red}{::}$aromatic(c_1$). The values are also not necessarily restricted to $(0,1)$, and can thus naturally represent numerical features, such as $\scalar{6}$\textcolor{red}{::}$atomicNumber(c_1)$ or $\scalar{2.35}$\textcolor{red}{::}$ionEnergy(c_1,level_2$). Finally the values are not necessarily restricted to scalars, and can thus {have the} form {of} feature vectors (tensors), such as $\scalar{[1.0, -7, \dots, 3.14]}$\textcolor{red}{::}$features(c_1)$.

{Ground facts in examples} are also not restricted to unary predicates, and can thus describe not only properties of individual objects, but values of arbitrary relational properties. For example, one can {assign} feature values to edges in graphs, such as describing a bond between two atoms $\scalar{[2.7, -1]}$\textcolor{red}{::}$bond(c_1,o_2)$.

{T}here is no syntactical restriction {on} how these representations can be mixed together, and one can thus select which parts of the data are better modelled with (sub-symbolic) distributed numerical representations, and which parts yield themselves to be represented by purely logical means, and move continuously along this dimension as needed.

\paragraph{Query:} Queries ($q$) (Sec.~\ref{sec:back:logic:prog}) represent the classification labels or regression targets associated with an example for supervised learning. They again utilize the same weighted fact representation such as $\scalar{1}$\textcolor{red}{::}$class$ or $\scalar{4.7}$\textcolor{red}{::}$target(c_1)$. Note that the target queries again do not have to be unary, and one can thus use the same format for different tasks. {For} example, for knowledge graph completion, we would use queries such as $\scalar{1.0}$\textcolor{red}{::}$coworker(alice,bob)$.

\subsubsection{Learning Template}

The weighted logic programs written in LRNNs are then often referred to as \textit{templates}. Syntactically, a learning template $\mathcal{T}$ is a set of weighted rules $\mathcal{T} = \{\alpha_i, \{W^{\alpha_i}_j\}\} = \{(W^i, c) \leftarrow (W_1^i, b_1), \dots, (W_k^i, b_k)\}$ where each $\alpha_i$ is a definite clause and each $W_j$ is some real-valued tensor, i.e. expressions of the form
\begin{centeredornot}
\begin{lstlisting}[mathescape=true]
$\weight{W^1}$ :: h$_1^1$($\dots$) $\impl$ $\weight{W^1_{1}}$ : b$^1_1$($\dots$) $\conj$ ... $\conj\weight{W^1_j}$ : b$^1_i$($\dots$).
$\weight{W^2}$ :: h$_1^2$($\dots$) $\impl$ $\weight{W^2_{1}}$ : b$^2_1$($\dots$) $\conj$ ... $\conj\weight{W^2_k}$ : b$^2_j$($\dots$).
$\dots$
$\weight{W^n}$ :: h$_p^q$($\dots$) $\impl$ $\weight{W^n_{1}}$ : b$^n_1$($\dots$) $\conj$ ... $\conj\weight{W^n_l}$ : b$^n_k$($\dots$).
\end{lstlisting}
\end{centeredornot}
where h$_i^j$'s and b$_i^j$'s are predicates forming positive, not necessarily different, literals, and $\weight{W_i^j}$'s are the associated tensors (also possibly reused in different places).





The template constitutes roughly what neural \textit{architecture} means in deep learning\footnote{We deliberately refrain from using the common term of neural ``model'', since a single template can have multiple logical (and neural) models.} -- i.e. it does not (necessarily) encode a particular model or knowledge of the problem, but rather a generic mode of computation.

\begin{example}
\label{ex:template}
Consider a simple template for learning with molecular data, encoding a generic idea that the representation of a (chemical) atom (e.g. $h(h_1)$) is dependent on the representation of the atoms adjacent to it. Given that a molecule can be represented by describing the contained atoms (e.g. $a(h_1)$) and bonds between them (e.g. $b(h_1,o_1)$), we can intuitively encode this idea by a following rule
\begin{centeredornot}
\begin{lstlisting}[mathescape=true]
$\weight{W_{h_1}}$ :: h(X) $\impl$ $\weight{W_a}$ : a(Y)$\conj\weight{W_b}$ : b(X,Y).
\end{lstlisting}
\end{centeredornot}
Moreover, one might be interested in using the representation of all atoms ($h(X)$) for deducing the representation of the whole molecule, for which we can write
\begin{centeredornot}
\begin{lstlisting}[mathescape=true]
$\weight{W_q}$ :: q $\impl$ $\weight{W_{h_2}}$ : h(X).
\end{lstlisting}
\end{centeredornot}
to derive a single ground query atom ($q$), which can be associated with the learning target of the whole molecule. The concrete semantics of this template then follows in the next section.
\end{example}

\subsection{Semantics: Computational Graphs Defined by LRNNs}
\label{sec:semantics}

To {explain} the correspondence between a {relational} template $\mathcal{T}$ and a ``neural architecture'', we {now describe} the mapping {that takes the template and a given example description, consisting of ground facts,} and {produces} a standard {neural model}. {Here, ``standard neural model'' refers to a} specific differentiable computational graph.

First, let $\mathcal{N}_l$ be the set of rules and facts obtained from the template and a learning example $\mathcal{N}_l = \mathcal{T} \cup E_l$ by removing all the tensor weights. For instance, if we had a weighted rule $\scalar{W}$\textcolor{red}{::}$h~\impl~\scalar{W_1}$\textcolor{red}{:}$b_1 \conj \scalar{W_2}$\textcolor{red}{:}$b_2$ , we would obtain $h~\impl~b_1\conj b_2$. {Then} we {construct} the least Herbrand model $\overline{\mathcal{N}_l}$ of {$\mathcal{N}_{l}$}, which can be done using standard theorem proving techniques. One simple option is to employ a bottom-up grounding strategy by repeated application of the immediate consequence operator (Section~\ref{sec:back:logic})\footnote{Another option is backward-chaining of the rules back from the associated query atom ($q$) through $\mathcal{T}$ into $E_l$. Note that this choice is purely technical and, following proper logical inference in both cases, does not affect the resulting logical (or neural) model.}. For consequent neural learning, the target query atom $q$ associated with $E_l$ must be logically entailed by $\mathcal{N}_{l}$, i.e. present in $\overline{\mathcal{N}_l}$\footnote{Otherwise it is automatically considered false (or having a default value) via CWA.}.

Having the least Herbrand model $\overline{\mathcal{N}_l}$ containing $q$, we can directly construct a neural computational graph $G_l$. Intuitively, the structure of the graph contains all the logical derivations of the target query literal $q$ from the example evidence ${E}_l$ through the template $\mathcal{T}$. 
Now {we} formally define the transformation mapping from $\mathcal{N}_l$ to a computational graph:

\begin{itemize}
\item For each weighted ground fact $(V_i,e)$ occurring directly in ${E}_l$, there is a node $F_{(V_i,e)}$ in the computational graph, called a {\em fact node}.
\item For each ground atom $h$ occurring in $\overline{\mathcal{N}} \setminus {E}_l$, there is a node $A_{h}$ in {the computational graph}, called an {\em atom node}.
\item For every rule $c \leftarrow b_1 \wedge \dots \wedge b_k \in {\mathcal{T}}$ and every grounding substitution $c\theta = h \in \overline{\mathcal{N}}$, there is a node $\textit{G}_{(c \leftarrow b_1 \wedge \dots \wedge b_k)}^{c\theta=h}$ in {the computational graph}, called an {\em aggregation node}.
\item For every \textit{ground} rule $\alpha_i\theta = (c\theta \leftarrow b_1\theta \wedge \dots \wedge b_k\theta) \in \overline{\mathcal{N}}$, there is a node $R_{(c\theta \leftarrow b_1\theta \wedge \dots \wedge b_k\theta)}$ in {the computational graph}, called a {\em rule node}.
\end{itemize}

\begin{table*}[t]
    \caption{Correspondence between the logical ground model and computation graph.}
    \label{overviewNNconstruction}
    \centering
    \begin{tabular}{llllc}
        Logical construct &     Type of node      & Notation  \\
        \hline
        Ground atom $h$ &   Atom node         &  $A_h$ \\
        Ground fact $h$ &   Fact node          & $F_{(h,\Vec{w})}$   \\
        Ground rule's $\alpha\theta$ body &   Rule node  &  $R_{(W_0^c c\theta \leftarrow W_1^{\alpha} b_1\theta\wedge\dots\wedge W_k^{\alpha} b_k\theta)}^{c\theta}$        \\
        Rule's $\alpha$ ground head $h$ &    Aggregation node  &  $\textit{G}_{(W_0^c c \leftarrow W_1^{\alpha} b_1 \wedge \dots \wedge W_k^{\alpha} b_k)}^{h=c\theta_i}$        \\
    \end{tabular}
\end{table*}



An overview of the correspondence between the logical and the neural model, together with the used notation is reviewed in Table~\ref{overviewNNconstruction}. 

The nodes {of the computational graph that we defined above} are then interconnected so as to follow the derivation of the {logical facts} by the immediate consequence operator starting from ${E}_l$, i.e. starting from the \textit{fact nodes} $F_{(V_i,e)}$ which have no antecedent inputs in {the computational graph} and simply output their associated values $out(F_{(V_i,e)}) = V_i$. The fact nodes are then connected into \textit{rule nodes} $R_{\alpha\theta}$, particularly a node $F_{(V_i,e)}$ will be connected into \textit{every} node $R_{\alpha\theta} = R_{(c\theta \leftarrow b_1\theta \wedge \dots \wedge b_k\theta)}$ where $e=b_i\theta$ for some $i$. 
Having all the inputs, corresponding to the body literals of the associated ground rule, connected, the rule node will output a value calculated as
$$out(R_{\alpha\theta}) = g_{\wedge}\big(W_1^{\alpha}\cdot out(F_{(V_1,b_1\theta)}),\dots, W_k^{\alpha}\cdot out(F_{(V_i,b_k\theta)})\big).$$ 
The rule node's activation function $g_{\wedge}$ is up to user's choice. For scalar inputs, {it} can be for example set to mimic {conjunction} from Lukasiewicz logic, {as in our previous work}~\citep{sourek2018lifted}. However, one can also choose to ignore the {fuzzy}-logical interpretation and use completely distributed semantics and activations utilized commonly in deep learning{.} In {this} case the computation follows the common (matrix) calculus by firstly aggregating the node's input values into its activation value
$$\underset{(1 \times l)}{{act(R_{\alpha\theta})}} = \underset{(l \times n)}{W_1^\alpha} \cdot \underset{(1 \times n)}{{out(F_1)}} + \dots + \underset{(l \times m)}{W_j^\alpha} \cdot \underset{(1 \times m)}{{out(F_k)}},$$ 
followed by an element-wise application of any differentiable function, such as logistic sigmoid
$$\underset{(1 \times l)}{{out(R_{\alpha\theta})}} = \sigma(\underset{(1 \times l)}{{act(R_{\alpha\theta})}}) = \sigma\big(act(R\alpha\theta)_1), \dots, \sigma(act(R\alpha\theta)_l\big).$$

\begin{figure*}[t!]
\centering
\resizebox{1.0\textwidth}{!}{
	\input{img/tikz/template}
}
\caption{A simple LRNN template with 2 rules described in Example 1. Upon receiving 2 example molecules, 2 neural computation graphs get created, as prescribed by the semantics (Section~\ref{sec:semantics}).}
\label{fig:template}
\end{figure*}
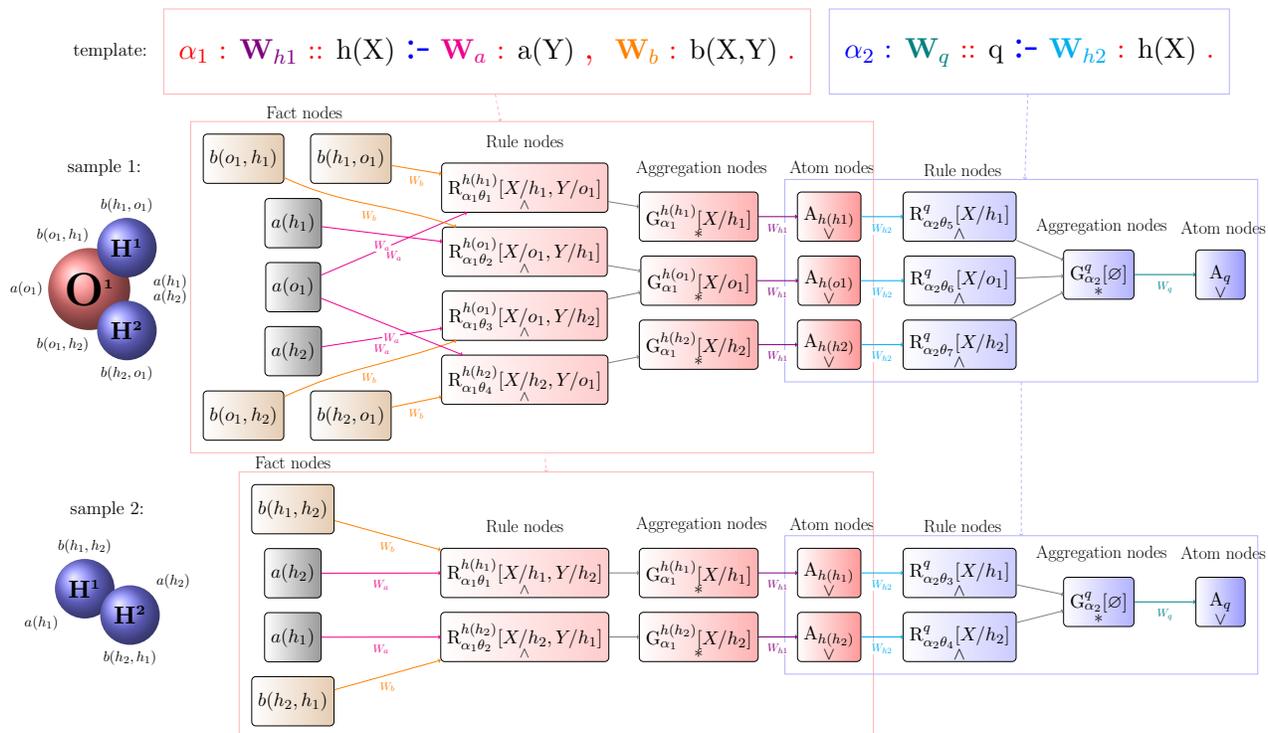

The rule nodes are then connected into \textit{aggregation nodes}. Particularly, a rule node $R_{(c\theta \leftarrow b_1\theta \wedge \dots \wedge b_k\theta)}$ is connected into \textit{the} aggregation node ${G}_{(c \leftarrow b_1 \wedge \dots \wedge b_k)}^{c\theta=h}$ {that corresponds} to the same ground head literal ${c\theta}$. Having all the inputs, corresponding to different grounding substitutions $\theta_i$ of the rule $c \leftarrow (b_1 \wedge \dots \wedge b_k)$ with the same ground head $h = c\theta_1 = \dots = c\theta_q$, connected, the aggregation node will output {the} value  
$$out(G_{\alpha}^c\theta=h) = g_*\big(out(R_{\alpha\theta_1}^{c\theta_1=h}),\dots,out(R_{\alpha\theta_q}^{c\theta_q=h})\big).$$
where $g_*$ is some aggregation function, such as $avg$ or $max$.
The aggregation nodes effectively aggregate all the different ways by which a literal $h$ can be derived from a single rule $\alpha$. Their semantic intuitively corresponds to a certain quantification over free variables appearing solely in the rule's $\alpha$ body. The aggregation $g_*$ is then applied in each dimension of the input values as
\begin{multline*}
\underset{1 \times l}{{out(G_{\alpha}^h)}} = g_*(\underset{1 \times l}{out(R_1)}, \dots, \underset{1 \times l}{out(R_q)}) = \Big( g_*\big(out(R_1)^1, \dots, out(R_q)^1\big), \dots, \\
\dots, g_*\big(out(R_1)^l, \dots, out(R_q)^l\big) \Big).    
\end{multline*}

The aggregation nodes are then connected into \textit{atom nodes}. {In particular}, {an} aggregation node $G_{\alpha}^h$ will be connected into \textit{the} atom node $A_{h}$ {that is} associated with the same atom $h$. The inputs of the atom node represent all the possible rules $\alpha_i$ through which the same atom $h$ can be derived. Having them all connected, $A_{h}$ will output {the} value
$$out(A_{h}) = g_{\vee}\big(W_1^c \cdot out(\textit{G}_{\alpha_1}^{h}), \dots, W_m^c \cdot out(\textit{G}_{\alpha_m}^{h})\big).$$ 
Apart from the choice of activation function $g_{\vee}$, the computation of the atom node's output follows exactly the same scheme as for the rule nodes. 

Finally, the atom nodes are connected into rule nodes in exactly the same fashion as fact nodes, i.e. $A_{h}$ will be connected into \textit{every} $R_{(c\theta \leftarrow b_1\theta \wedge \dots \wedge b_k\theta)}$ where $h=b_i\theta$ for some $i$, and the whole process continues recursively.

\begin{example}
\label{ex:ground}
Let us follow up on the Example 1 by extending the described template with two example molecules of hydrogen and water. The template will then be used to dynamically unfold two computation graphs, one for each molecule, as depicted in Figure~\ref{fig:template}. Note that the computation graphs have different structures, following from the different Herbrand models derived from each molecule's facts, but share parameters in a scheme determined by the lifted structure of the joint template.
\end{example}

\section{Examples of Common Neural Architectures}
\label{sec:examples}

We demonstrate flexibility of the declarative LRNN templating, stemming from the abstraction power of Datalog, by encoding a variety of common neural architectures in very simple differentiable logic programs. For completeness, we start from simple neural models, where the advantages of templating are not so apparent, but continue to advanced deep learning architectures, where the expressiveness of relational templating stands out more clearly. Note that all templates in this paper are actual programs that can be run and trained with the LRNN interpreter.


\subsection{Feed-forward Neural Networks}
\label{sec:MLPs}



Multi-layer perceptrons (MLPs) form the most simple case where the weighted logic template is restricted to propositional clauses, and its single Herbrand model thus directly corresponds to a single neural model (Section~\ref{sec:semantics}). In this setting, the input example information can thus be encoded merely in the \textit{values} of their associated tensors, which is the standard deep learning scenario. In the vector form, we can associate each example $E_i$ with a fact proposition $\weight{[v_1^i,\dots,v_n^i]}$\textcolor{red}{::}$features^{(0)}$, forming the input ($0$-th) node of the neural model. Each example is further associated with a query $\weight{v_q^i}$\textcolor{red}{::}$q$.

{In particular, an} MLP with 3 layers, i.e. input layer$^{(0)}$, 1 hidden layer$^{(1)}$, and output layer$^{(2)}$, with the corresponding weight matrices $[\weight{\underset{m \times n}{W}^{(1)}}, \weight{\underset{1 \times m}{W}^{(2)}}]$ can be directly modelled with the following rule

\begin{centeredornot}
\begin{lstlisting}
$\weight{\underset{1 \times m}{W}^{(2)}}$ :: q$^{(2)}$ $\impl$ $\weight{\underset{m \times n}{W}^{(1)}}$ : features$^{(0)}$.
\end{lstlisting}
\end{centeredornot}
Naturally, we can extend {it} to a deeper MLP by stacking more rules as
\begin{centeredornot}
\begin{lstlisting}
$\weight{\underset{r \times m}{W}^{(2)}}$ :: hidden$^{(2)}$ $\impl$ $\weight{\underset{m \times n}{W}^{(1)}}$ : features$^{(0)}$.
$\dots$
$\weight{\underset{1 \times s}{W}^{(k)}}$ :: q$^{(k)}$ $\impl$ $\weight{\underset{s \times r}{W}^{(k-1)}}$ : hidden$^{(k-2)}$.
\end{lstlisting}
\end{centeredornot}
Once the template gets transformed into the corresponding neural model (Sec.~\ref{sec:semantics}), its computation graph will consist of a linear chain of nodes (Sec.~\ref{sec:back:mlp}) corresponding to standard fully-connected layers $1,\dots,k$ with associated weight matrices $[{W}^{(1)},{W}^{(2)}\dots,{W}^{(k)}]$, and activation functions of user's {choice}.

\begin{figure*}[t!]
\centering
\resizebox{1.0\textwidth}{!}{
	\input{img/tikz/MLPs}
}
\caption{Demonstration of the pruning technique on a sample MLP model unfolded from a 2-rule template of  $\protect\alpha_1 = \protect\wc{orange}{W_1}\textcolor{red}{::} h_1 \protect\impl f.$ and $\protect\alpha_2 = \protect\wc{violet}{W_2}\textcolor{red}{::} h_2 \protect\impl h_1$.}
\label{fig:pruning}
\end{figure*}
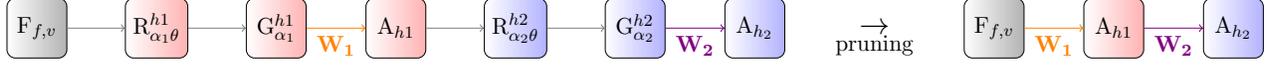

Note that not all the weights need to be specified, and one can thus also write, e.g., either of
\begin{centeredornot}
\begin{lstlisting}
$\weight{W}$ :: h$^{(2)}$ $\impl$ $h^{(0)}$.                 h$^{(2)}$ $\impl$ $\weight{W}$ : $\textit{h}^{(0)}$.
\end{lstlisting}
\end{centeredornot}
While each of these rules still encodes in essence a 3-layer MLP, either only the hidden (right) or only the output (left) layer will carry learnable parameters, respectively. Moreover, following the exact semantics (Sec~\ref{sec:semantics}) for neural model creation, an aggregation node will be created on top of a rule node, representing the hidden layer. Since there is no need for aggregation in MLPs, i.e. only a single rule node ever gets created from each propositional rule, this introduces unnecessary operations in the graph. Since such nodes arguably do not improve learning of the model, we \textit{prune} them out, as depicted in Figure~\ref{fig:pruning}. The technique is further described in more detail in Section~\ref{sec:app:pruning}. Note that we assume application of pruning, where applicable, in the remaining examples described in this paper.





\subsubsection{Knowledge-based Artificial Neural Networks}
{The direct correspondence between a propositional program and a neural network has been successfully exploited in a number of previous works, particularly the original Knowledge-Based Artificial Neural Networks (KBANN)~\citep{KBANN}}, and LRNNs can be seen as a direct extension of KBANN into relational setting.
To emulate the KBANN inference and learning, we simply fall back to scalar representation of features, e.g. $\scalar{0}$\textcolor{red}{::}$rain, \scalar{0.6}$\textcolor{red}{::}$wet, \scalar{0.8}$\textcolor{red}{::}$sunny$, and consider a propositional template encoding some background domain knowledge, such as $sprinkle~\impl~wet~\conj~sunny$. One then needs to choose a set of proper activation functions based on desired multi-valued logic semantics, e.g. the Lukasiewicz's fuzzy operators~\citep{KBANN,sourek2018lifted}. Note that, choosing proper fuzzy logic activations, this still covers standard logical inference as a special case with the use of binary fact values.

\subsection{Convolutional Neural Networks}
\label{sec:CNNs}
The CNNs can no longer be represented with a propositional template. To emulate the additional parts w.r.t. the MLPs, i.e. the convolutional filters and pooling (Sec.~\ref{sec:back:neural}), we need to move to \textit{relational} rules (Sec.~\ref{sec:back:logic}). Note that there is a natural, close relationship between convolutions and relational rules (or relational patterns in general), where the point of both is to exploit symmetries in data. Moreover, the point of both the aggregation nodes and the pooling layers is to enforce certain transformation invariance. Let us demonstrate this relationship with the following example.

For clarity of presentation, consider a simplistic one-dimensional ``image'' consisting of 5 pixels $i=1,...,5$. While the regular grid structure of the image pixels is inherently assumed in CNN, we will need to encode it explicitly. Considering the 1-dimensional case, it is enough to define a linear ordering of the pixels such as $next(1,2),\dots,next(4,5)$. The (gray-scale) value $v_i$ of each pixel $i$ can then be encoded by a corresponding weighted fact $\scalar{v_i}\dw f(i)$. Next we encode a convolution filter of size $[1,3]$, i.e. vector which combines the values of each three ([\textcolor{orange}{l}eft,\textcolor{cyan}{m}iddle,\textcolor{violet}{r}ight]) consecutive pixels, and a (max/avg)-pooling layer that aggregates all the resulting values. This computation can be encoded using the following template

\begin{figure*}[t!]
\centering
\resizebox{1.0\textwidth}{!}{
	\input{img/network/CNN}
}
\caption{Left: core part of a standard CNN architecture with sparse layer composed of sequential applications of a convolutional filter (h), creating a feature-map layer, followed by a pooling operator. Right: the corresponding computation graph derived from a LRNN template.}
\label{fig:CNN}
\end{figure*}
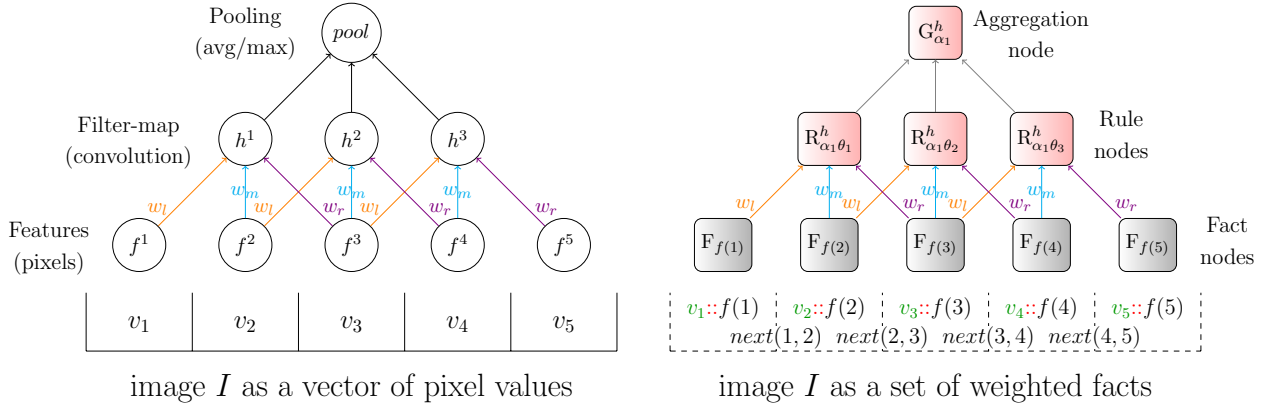

\begin{centeredornot}
\begin{lstlisting}
h $\impl$ $\textcolor{orange}{w_l}$: f(A)$\conj~\textcolor{cyan}{w_m}$: f(B)$\conj~\textcolor{violet}{w_r}$: f(C)$\conj$next(A,B)$\conj$next(B,C).
\end{lstlisting}
\end{centeredornot}
A visualization of the CNN and the corresponding computation graph derived from the logic model of the template presented with some example pixel values $[v_1,\dots,v_5]$ is shown in Fig~\ref{fig:CNN}.


While this does not seem like a convenient way to represent learning with CNNs from images, the important insight is that convolutions in neural networks correspond to weighted relational rules (patterns). The efficiency of normal CNN encoding is due to the inherent assumptions that are present in CNNs w.r.t.\ topology of their application domain, i.e. grids of pixel values, and similarly complete, ordered structures. 
While with LRNNs we need to state all these assumptions explicitly, it also means that we are not restricted to them -- an advantage which will become clearer in the subsequent sections.

\subsection{Recursive and Recurrent Neural Networks}
\label{sec:RNNs}

A \textit{recursive} network also exploits the principle of convolution, however the input is {no longer} a grid but a regular tree of an unknown structure. This prevents us from creating computation schemes customized to a specific structure, as in the CNNs. Instead, we need to resort to a general convolutional pattern that can be applied over any $k$-regular tree.

For that purpose, we again utilize the expressiveness of relational logic. Firstly, we encode the $k$-regular tree structure itself by providing a fact connecting each parent node in the tree to its child-nodes, i.e. $parent(node^{i+1}_j,node^{i}_l,\dots,node^{i}_{l+k})$. Secondly, we associate all the leaf nodes in the tree with their embedding vectors $\weight{[v_1^i,\dots,v_n^i]}\dw\dw n(\textit{leaf}_i)$. Finally, a single relational rule can then be used to encode the recursive composition of representations in the, for instance $3$-regular, tree as
\begin{centeredornot}
\begin{lstlisting}
n(P) $\impl$ $\wc{orange}{W_1}$:$$n(C$_1$)$\conj\wc{cyan}{W_2}$:$$n(C$_3$)$\conj\wc{violet}{W_3}$:$$n(C$_3$)$\conj$parent(P,C$_1$,C$_2$,C$_3$).
\end{lstlisting}
\end{centeredornot}
\noindent which directly forms the whole learning template. Given a particular example tree, this rule translates to a computation graph recursively combining the children node representations ($n(C)$) into respective parent node representations, until the root node is reached. The root node representation ($n(root)$) could then be fed into a standard MLP rule (Sec.~\ref{sec:MLPs}) to output the value for a given target query associated with the whole tree example.

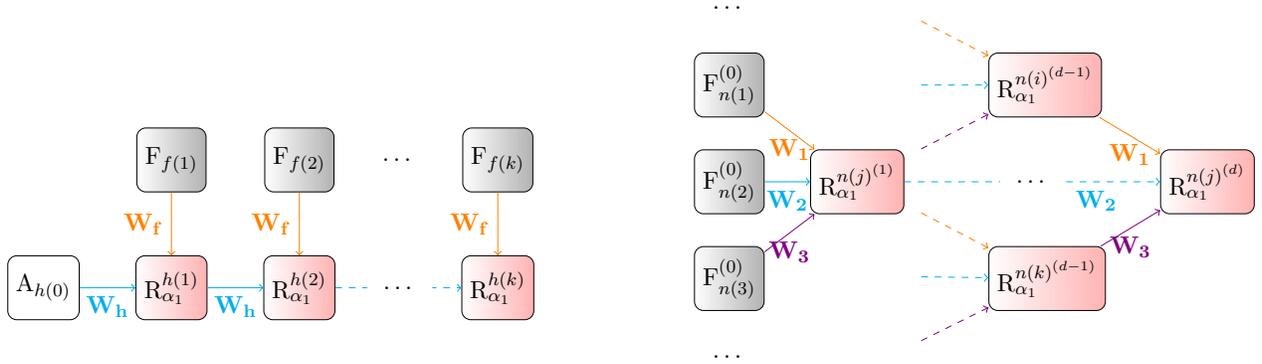
\begin{figure*}[t!]
\centering
\resizebox{1.0\textwidth}{!}{
	\begin{tikzpicture}
	  \node[scale=1] (rec1) at (0,0){\input{img/tikz/recurrent}};
	  \node[scale=1] (rec2) at (11,0.7){\input{img/tikz/recursive}};
	\end{tikzpicture}
}
\caption{Simple recurrent (left) and recursive (right) neural structures encoded through LRNNs.}
\label{fig:rec}
\end{figure*}

A simple \textit{recurrent} neural network unfolded over a linear (time) structure can then be modelled in a simpler manner, where only a single (vector) input is given at each step and a linear chain of \textit{hidden} nodes ($h(X)$) replaces the prescribed tree hierarchy. Assuming encoding of the linear example structure with predicate $next(X,Y)$ as before, such a model can then be written as
\begin{centeredornot}
\begin{lstlisting}
h(Y) $\impl$ $\wc{orange}{W_f}$:$$f(Y)$\conj\wc{cyan}{W_h}$:$$h(X)$\conj$next(X,Y).
\end{lstlisting}
\end{centeredornot}
The final hidden representation ($h(k)$) could then again be fed into a MLP for a whole sequence-level prediction. Neural architectures of both these templated models are displayed in Figure~\ref{fig:rec}.
\\\\
\section{Graph Neural Networks in LRNNs}
\label{sec:GNNs}

Graph (convolutional) Neural Networks (GNNs) (Sec.~\ref{sec:back:gnn}) can be seen as a generalization of the introduced neural architectures (Section~\ref{sec:examples}) to arbitrary graphs, for which they combine the ideas of latent representation learning (Sec.~\ref{sec:MLPs}), convolution (Sec.~\ref{sec:CNNs}), and dynamic model structure (Sec.~\ref{sec:RNNs}).

While modelling CNNs in the weighted logic formalism was somewhat cumbersome (because we had to explicitly represent the pixel grid), the encoding of GNNs is very straightforward. This is due to the underlying general graph representation with no additional assumptions of its structure, which yields itself very naturally to relational logic. The computation of the layer $i$ update in GNNs can then be represented by a single rule as follows

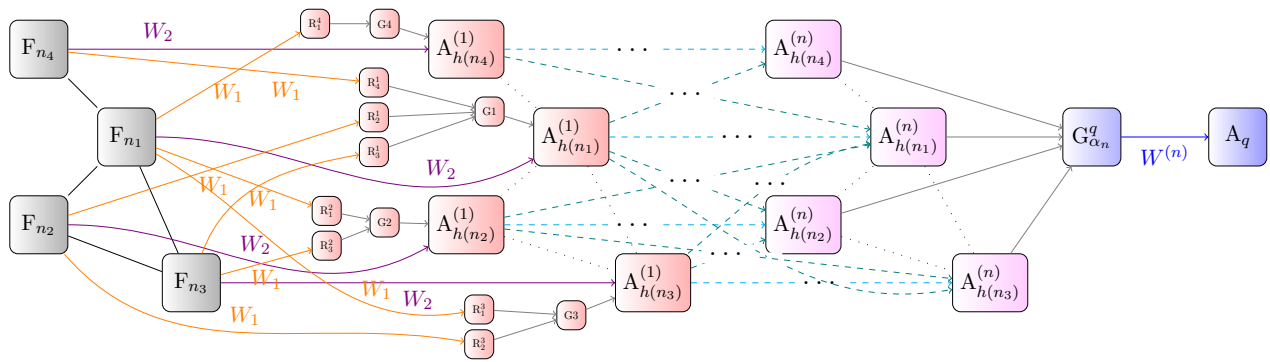
\begin{figure*}[t!]
\centering
\resizebox{1\textwidth}{!}{
\input{img/tikz/GNNs}
}
\caption{A computation graph of a sample (g-SAGE) GNN as encoded in LRNNs. Given an input graph of 4 (fact) nodes (F$_{n_1}\dots$F$_{n_4}$), neighbors of each node are firstly weighted and aggregated with rule and aggregation nodes, respectively (reduced in size in picture). The result is then combined with representation of the (central) node from the preceding layer, to form a new layer of 4 atom nodes, copying the structure of the input graph. After $n$ such layers, each with the same structure but different parameters, a global readout (aggregation) node aggregates all the node representations, passing to the final query (atom) node's transformation.}
\label{fig:GNNs}
\end{figure*}

\begin{centeredornot}
\begin{lstlisting}
$\weight{W^{(i)}}$:: $$h$^{(i)}$(V) $\impl$ h$^{(i-1)}$(U)$\conj$edge(V,U).
\end{lstlisting}
\end{centeredornot}
where $edge/2$ is the binary relation of the given input graphs. With the choice of activation functions as $g_* = avg, g_\wedge = ReLU$, this simple rule can be already used to model the popular Graph Convolutional Neural Networks (GCN)~\citep{kipf2016semi}\footnote{where the authors also denoted the rule as \textit{convolution}, since it forms a linear approximation of a localized spectral convolution~\citep{kipf2016semi}.}. The exact same rule (up to parameterization) is then used at each layer. For the final output query ($q$) representing the whole graph we simply aggregate representations of all the nodes as
\begin{centeredornot}
\begin{lstlisting}
$\weight{W^{(d)}}$:: q $\impl$ h$^{(d-1)}$(U).
\end{lstlisting}
\end{centeredornot}
A noticeable shortcoming of GCNs is that the representation of the ``central'' node (V) itself is not used in the representation update. While this can be done by extending the graph ($edge/2$) with self-loops, a novel\footnote{Note that, differently from GCN with self-loops, the central node is parameterized differently from the neighbors.} GNN model called GraphSAGE (g-SAGE)~\citep{hamilton2017inductive} was proposed to address this explicitly. To follow the architecture of g-SAGE, we thus split the template into 2 rules accordingly
\begin{centeredornot}
\begin{lstlisting}
$$h$^{(i)}$(V) $\impl$ $\wc{orange}{W_1^{(i)}}$: h$^{(i-1)}$(U)$\conj$edge(V,U).
$$h$^{(i)}$(V) $\impl$ $\wc{violet}{W_2^{(i)}}$: h$^{(i-1)}$(V).
\end{lstlisting}
\end{centeredornot}
and choose $g_\wedge = ReLU, g_* = max, g_\vee = identity$ for the very model (g-SAGE), the depiction of which can be seen in Figure~\ref{fig:GNNs}.

Another popular extension taken from neural architectures for image recognition are residual (skip) connections, where one effectively adds links to preceding layers at arbitrary depth (instead of just the preceding layer), i.e. we simply add one or more rules in the form
\begin{centeredornot}
\begin{lstlisting}
$\weight{W_{skip}^{(i)}}$::$$h$^{(i)}$(V) $\impl$ h$^{(i-skip)}$(V).
\end{lstlisting}
\end{centeredornot}
This technique is also used in the Graph Isomorphism Network (GIN)~\citep{xu2018powerful}, which is a theoretically substantiated GNN based on the expressive power of the Weisfeiler-Lehman test (WL)~\citep{weisfeiler2006construction}.
Firstly, the GIN model differs in that it adds residual connections from all the preceding layers to the final layer (which the authors refer to as ``jumping knowledge''~\citep{xu2018representation}).
Secondly, the particularity of GIN is to add a 2-layered MLP on top of each aggregation to harvest its universal approximation power. Particularly, update formula derived from the WL-correspondence~\citep{xu2018powerful} is
$$h^{(i)}(v) = MLP^{(i)}\big((1+\epsilon^{(i-1)}) \cdot h^{(i-1)}(v) + \sum_{u \in \mathcal{N}(v)} h^{(i-1)}(u)\big)$$
where $MLP$ is the 2-layered MLP (Sec.~\ref{sec:back:mlp}). To accommodate the extra MLP layer, we thus extend the template as follows~
\begin{centeredornot}
\begin{lstlisting}
mlp$_{tmp}^{(i)}$(V) $\impl$ h(U)$^{(i-1)}$$\conj$edge(V,U).
mlp$_{tmp}^{(i)}$(V) $\impl$  $\weight{(1+\epsilon^{(i-1)})}$ : h$^{(i-1)}$(V).
$\weight{W^{(i)}_{{2}}}$ :: h$^{(i)}$(V) $\impl$ $\weight{W^{(i)}_{{1}}}$: mlp$_{tmp}^{(i)}$(V).
\end{lstlisting}
\end{centeredornot}
Note that, considering that such a single rule actually already models a 2-layer\footnote{or 3-layer, depending on inclusion of the input layer in the count.} MLP (as described in Sec.~\ref{sec:MLPs}), a very similar computation can be carried out more simply with
\begin{centeredornot}
\begin{lstlisting}
$\weight{W_{2a}^{(i)}}$ :: h$^{(i)}$(V) $\impl$ $\weight{W_{1a}^{(i)}}$ : h(U)$^{(i-1)}$$\conj$edge(V,U).
$\weight{W_{2b}^{(i)}}$ :: h$^{(i)}$(V) $\impl$ $\weight{W_{1b}^{(i)}}$ : h$^{(i-1)}$(V).
\end{lstlisting}
\end{centeredornot}
corresponding to a GIN version without the special ${(1+\epsilon^{(i)})}$ coefficient, which the authors refer to as ``GIN-0''~\citep{xu2018powerful} and actually find performing better\footnote{We note there is a slight difference, where GIN-0 firstly aggregates the neighbors and weights the result, while this template aggregates the neighbors after weighting. Nevertheless we note that GNN authors often switch this order themselves, for instance GraphSAGE in~\citep{dwivedi2020benchmarking} performs weighting before aggregation, while it is vice-versa in~\citep{xu2018powerful}.}.
Finally they choose $g_*=sum$ as the function to aggregate the neighborhood representations. The authors proved the GIN model to belong to the most ``powerful'' class of GNN models, i.e. no other GNN model is more expressive than GIN, and demonstrated the GIN-0 model to provide state-of-the-art performance in various graph classification and completion tasks~\citep{xu2018powerful}.



\subsection{Extending GNNs}


While the GIN model presents the most ``powerful'' version of the basic GNN idea, there is a large number of ways in which the GNN approach can be extended. We discuss some of the direct, natural extensions in this subsection.

\subsubsection{Edge Representations}
Originally aimed at single-relation graphs, GNNs do not adequately utilize the information about the possibly different types of edges. While it is straightforward to associate edges with scalar weights in the adjacency matrix, instead of using just binary edge indicators~\citep{kipf2016semi}, extending to richer edge representations is not so direct, and has only been explored recently~\citep{kipf2018neural,gong2019exploiting,kim2019edge}.

In the templating approach, addressing edges is very simple, since we do not operate directly on the graph but on the ground logical model, where each edge ($edge(n_1,n_2)$) forms an \textit{atom} in exactly the same way as the actual nodes ($node(n_1)$) in the graph itself (similarly to an extra transformation introduced in line-GNNs~\citep{chen2017supervised}). We can thus directly associate edges corresponding to different relations with arbitrary features ($\weight{[v_1,\dots,v_n]}\textcolor{red}{::}~edge(n_1,n_2)$), learn their distributed representations, and predict their properties (or existence), just like GNNs do with the nodes. For basic learning with edge representations, there is no need to change anything in the previously introduced templates.
However, one might want to associate extra transformations for edge and node representation learning~\citep{gong2019exploiting}, in which case we would simply write
\begin{center}
\begin{tabular}{c}
\begin{lstlisting}
$\weight{W^{(i)}}$ :: h$^{(i)}$(V) $\impl$ h$$(U)$^{(i-1)}$$\conj$$\weight{W_{e}}$:$$edge$$(V,U).
\end{lstlisting}
\end{tabular}
\end{center}

A large number of structured data then come in the form of {multi-relational} graphs, where the edges can take on different types. A straightforward extension is to learn a separate node representation of the nodes for each of the relations, e.g. as
\begin{center}
\begin{tabular}{c}
\begin{lstlisting}
$\weight{W^{(i)}}$ :: h$_x^{(i)}$(V) $\impl$ h$_x$(U)$^{(i-1)}$$\conj$$\weight{W_{e}}$:$$edge$_{type=x}$(V,U).
\end{lstlisting}
\end{tabular}
\end{center}
and to choose from the different representations depending on context, such as in multi-sense word embeddings~\citep{li2015multi}, or simply directly combine~\citep{schlichtkrull2018modeling} these representations in the template.

\subsubsection{Heterogeneous Graphs}
\label{sec:hetero}
The majority of current GNNs then assume homogeneous graphs, and learning from heterogeneous graphs has just been marked as one of the future directions for GNNs~\citep{wu2020comprehensive}. In LRNNs, various heterogeneous graphs~\citep{wang2019heterogeneous} can be directly covered without any modification, since there is no restriction to the types of nodes and relations to be used in the same template (and so we do not have to e.g. split the graphs~\citep{zhu2019relation} or perform any extra operation~\citep{liu2018heterogeneous} for such a task).
In the context of heterogeneous information networks, a similar ``templating'' idea has already become popular as defining \textit{``meta-paths''}~\citep{dong2017metapath2vec,huang2017heterogeneous}, which can be directly covered by a single LRNN rule and, importantly, differentiated through.

We can further represent the relations as actual objects to be operated by logical means, by reifying them into logical constants as
\begin{center}
\begin{tabular}{c}
\begin{lstlisting}
$\weight{W_1^{(i)}}$ :: h$^{(i)}$(V) $\impl$ h$$(U)$^{(i-1)}\conj$h$$(E)$^{(i-1)}$$\conj$edge(V,U,E).
\end{lstlisting}
\end{tabular}
\end{center}
where variable E represents the edge object and h(E) is its hidden representation. The learned embeddings of the nodes and relations can then be directly used for predicting triplets of\\ ({O}$bject$,{R}$elation$,{S}$ubject$) in KBC, again with a simple template extension, e.g. for an MLP-based KBE~\citep{dong2014knowledge}, as
\begin{center}
\begin{tabular}{c}
\begin{lstlisting}
$\weight{W}$ :: edge(O,R,S) $\impl$ $\weight{W_o}$:$$h(O)$\conj\weight{W_r}$:$$h(R)$\conj\weight{W_s}$:$$h(S).
\end{lstlisting}
\end{tabular}
\end{center}

\subsubsection{Hypergraphs}
\label{sec:hypergraphs}
Naturally, the GNN idea can be extended to hypergraphs, too, as was recently also proposed~\citep{feng2019hypergraph}. While extending to hypergraphs from the adjacency matrix form used for simple graphs can be somewhat cumbersome, in the relational Datalog, hypergraphs are first-class citizens, so we can just directly write
\begin{center}
\begin{tabular}{c}
\begin{lstlisting}
$\weight{W_1^{(i)}}$ :: h$^{(i)}$(U$_1$) $\impl$ h(U$_1$)$^{(i-1)}$ $\conj\dots\conj$h(U$_n)^{(i-1)}$ $\conj$edge($U_1,\dots,U_n$).
$\dots$
$\weight{W_1^{(i)}}$ :: h$^{(i)}$(U$_n$) $\impl$ h(U$_1$)$^{(i-1)}$ $\conj\dots\conj$h(U$_n)^{(i-1)}$ $\conj$edge($U_1,\dots,U_n$).
\end{lstlisting}
\end{tabular}
\end{center}
and possibly combine with all the other extensions.

There are many other simple ways in which GNNs can be extended towards higher expressiveness and there is a wide variety of emerging works in this area. While reaching beyond the standard, single adjacency matrix format, each of the novel extensions typically requires extra transformations (and libraries) to create their necessary intermediate representations~\citep{chen2017supervised,dong2017metapath2vec}. Many of these extensions are often deemed complex from the graph (GNN) point of view, but are rather trivial template modifications with LRNNs, as indicated in the preceding examples (and following in Sec.~\ref{sec:beyond}). This is due to the adopted \textit{declarative} relational abstraction, as opposed to the procedural manipulations on ground graphs, defined often on a per basis.
On the other hand we note that LRNNs currently cannot cover recent non-isotropic GNNs~\citep{dwivedi2020benchmarking} with computation constructs such as attention, gating, or LSTMs~\citep{wu2020comprehensive}. While such construct could be included on an ad-hoc basis, they do not yield themselves naturally to the LRNN semantics.








\subsection{Beyond GNN architectures}
\label{sec:beyond}

While we discussed possible ways for direct extensions of GNNs, there are more substantial alterations that break beyond the core principles of GNNs. One of them is the ``message passing'' idea, where the nodes are restricted to ``communicate'' with neighbors through the existing edges (WL label propagation~\citep{weisfeiler2006construction}). Obviously, there is no such restriction in LRNNs, and we can design templates for arbitrary message passing schemes, corresponding to more complex and expressive convolutional filters. For instance, consider a simple extension beyond the immediate neighborhood aggregation by defining edges as weighted paths of length 2 (introduced as ``soft edges'' in~\citep{sourek2018lifted}):

\begin{center}
\begin{tabular}{c}
\begin{lstlisting}
$\weight{W}$:: edge2(U,W) $\impl$ $\weight{W_1}$:$$edge(U,V)$\conj\weight{W_2}$:$$edge(V,W).
\end{lstlisting}
\end{tabular}
\end{center}

We can also easily compose the edges into small subgraph patterns of interest (also known as ``graphlets'' or ``motifs'' used in, e.g., social network analysis~\citep{vsourek2013predicting}), such as triangles and other small cliques (alternatively conveniently representable by the hyper-edges (Sec.~\ref{sec:hypergraphs})), and operate on the level of these instead:

\begin{center}
\begin{tabular}{c}
\begin{lstlisting}
$\weight{W}$:: node(U,V,W) $\impl$ $\weight{W_1}$:$$edge(U,V)$\conj\weight{W_2}$:$$edge(V,W)$\conj\weight{W_3}$:$$edge(W,U).
\end{lstlisting}
\end{tabular}
\end{center}

Since both nodes and edges can be treated uniformly as logic atoms, we can easily alter the GNN idea to hierarchically propagate latent representations of the edges, too. In other words, each edge can aggregate representations of ``adjacent'' edges from previous layers:

\begin{center}
\begin{tabular}{c}
\begin{lstlisting}
$\weight{W}$:: h$_{edge}^{(i)}$(E) $\impl$ $\weight{W_{F}}$:$$h$_{edge}^{(i-1)}$(F)$\conj\weight{W_{U,V}}$:$$edge(U,V,E)$\conj\weight{W_{\small V,W}}$:$$edge(V,W,F).
\end{lstlisting}
\end{tabular}
\end{center}
\noindent
Naturally, this can be further combined with the standard learning of the latent node representations (as we do in experiments in Sec.~\ref{sec:experiments}).

Moreover, the messages do not have to spread homogeneously through the graph and a custom logic can drive the diffusion scheme. This can be, for instance, naturally put to work in the heterogeneous graph environments (Sec.~\ref{sec:hetero}) with explicit types, which can then be used to control communication and representation learning of the nodes:
\begin{center}
\begin{tabular}{c}
\begin{lstlisting}
$\weight{W}$:: h$^{(i)}$(X) $\impl$ h$^{(i-1)}$(Y)$\conj$edge(X,Y,E)$\conj$type(E,type$^e_{1}$).
\end{lstlisting}
\end{tabular}
\end{center}

Besides being able to represented the \textit{types} explicitly as objects (as opposed to the vector embeddings), we can actually induce new types, for instance into latent hierarchical categories (such as in~\citep{vsourek2016learning}):
\begin{center}
\begin{tabular}{c}
\begin{lstlisting}
isa(edge$_1$,type$^e_{1}$).
$\dots$
$\weight{W^{(1)}}$:: isa(supertype$^{(1)}_{e}$,type$^e_{1}$).
$\weight{W^{(k)}}$:: isa(A,C) $\impl$ $\weight{W^{(k-1)}_1}$: isa(A,B)$\conj\weight{W^{(k-1)}_2}$: isa(B,C).
\end{lstlisting}
\end{tabular}
\end{center}

Importantly, there is no need to directly follow the input graph structure in each layer. We can completely abstract away from the graph representation in the subsequent layers and reason on the level of the newly invented, logically derived, entities, such as, e.g., the various graphlets, latent types, and their combinations:

\begin{center}
\begin{tabular}{c}
\begin{lstlisting}
$\weight{W}$:: node$_{motif}^{(1)}(T_1,T_2,T_3)~\impl~\weight{W_1}$:node(X)$\conj\weight{W_2}$:$$node(Y)$\conj\weight{W_3}$:$$node(Z)$\conj$
                $\weight{W_4}$:$$type(X,T1)$\conj\weight{W_5}$:$$type(Y,T2)$\conj\weight{W_6}$:$$type(Z,T3)$\conj$
                                edge(X,Y)$\conj$edge(Y,Z)$\conj$edge(X,Z).
\end{lstlisting}
\end{tabular}
\end{center}

Finally, the models can be directly extended with external relational background knowledge. Note that such knowledge can be specified declaratively, with the same expressiveness as the templates themselves, since they are consequently simply merged together, for instance:

\begin{center}
\begin{tabular}{c}
\begin{lstlisting}
ring$_6(A,\dots,F)~\impl$ $\weight{V_{e1}}$:$$edge(A,B)$\conj\dots\conj\weight{V_{e6}}$:$$edge(F,A)$\conj$
                    $\weight{V_{n1}}$node(A)$\conj\dots\conj\weight{V_{n6}}$node(F).
$\weight{W}$:: node$^{(n)}$(X) $\impl$ $\weight{V_1}$: ring$_6(X,\dots,F$).
\end{lstlisting}
\end{tabular}
\end{center}
Note that this is very different from the standard GNNs, where one can only input ground information, in the form of numerical feature vectors along with the actual nodes (and possibly edges). Nevertheless this does not mean that LRNNs cannot work with numerical representations. On the contrary, besides the standard neural means, one can also directly interact with it by the logical means, e.g. by arithmetic predicates to define learnable numerical transformations (such as in~\citep{sourek2018lifted}) over some given (or learned) node similarities:

\begin{center}
\begin{tabular}{c}
\begin{lstlisting}
$\weight{W}$:: edge$_{sim}$(N$_1$,N$_2$) $\impl$ similar(N$_1$,N$_2$,S$_{im}$)$\conj\weight{W_{0.3}}$:$\geq$(S$_{im}$,0.3).
\end{lstlisting}
\end{tabular}
\end{center}





    
        






\section{Experiments}
\label{sec:experiments}

The preceding examples were meant to demonstrate high expressiveness and encoding efficiency of declarative LRNN templating. The main purpose of the experiments is to assess correctness and efficiency of the actual learning. For that purpose, we select GNNs as the most general and flexible of the commonly used neural architectures, since they encompass building blocks of all the other introduced architectures. Given the focus on GNNs, we compare against two most popular\footnote{with, as of date, PyG having 7.3K stars and DGL having 4.7K stars on Github, respectively.} GNN frameworks of Pytorch Geometric (PyG)~\citep{fey2019fast} and Deep Graph Library (DGL)~\citep{wang2019deep}. Both these frameworks contain reference implementations of many popular GNN models, which makes them ideal for such a comparison. Note also that both these frameworks are highly contemporary and were specifically designed and optimized for creation and training of GNNs.
For clarity of presentation, we restrict ourselves to a single task of structure property prediction, but perform experiments across a large number of datasets.

\subsection{Modern GNN frameworks}

While popular deep learning frameworks such as TensorFlow or Pytorch provide ways for efficient acceleration of standard neural architectures such as MLPs and CNNs, implementing GNNs is more challenging due to the irregular, dynamic, and sparse structure of the input graph data. Nevertheless, following the success of vectorization of the classic neural architectures, both PyG and DGL adopt the standard (sparse) tensor representation of all the data to leverage vectorized operations upon these. This includes the graphs themselves, which are then represented by their sparse adjacency matrices $G_i^i$. Further, each node index $i$ can be associated with a feature vector (${[f_1,\dots,f_j]}_i$) through an additional matrix $F_i^j$ associated with each input graph.

Following the standard procedural differentiable programming paradigm, both frameworks then represent model computations explicitly through a predefined graph of tensor transformations applied directly to the input graph matrices, creating an updated feature tensor $F_i^{j^{(k)}}$ at each step $k$. The same tensor transformations are then applied to each input graph.

Both frameworks are based on similar ideas of message passing between the nodes (neighborhood aggregation) and its respective acceleration through optimized sparse tensor operations and batching~(gather-scatter). DGL then seems to support a wider range of operations (and backends), with high-level optimizations directed towards larger scale data and models (and a larger overhead), while PyG utilizes more efficient low-level optimizations stemming from its tighter integration with Pytorch.

\subsection{Model and Training Correctness}

Firstly, we evaluate correctness of the templated GNN architectures via correspondence to their reference implementations in PyG and DGL. For this we select some of the most popular GNN models introduced in previous chapters, particularly the original GCN~\citep{kipf2016semi}, highly used GraphSAGE (g-SAGE)~\citep{hamilton2017inductive} and the ``most powerful'' GIN~\citep{xu2018powerful} (particularly GIN-0). Each of the models comes with a slightly different aggregate-combine scheme and particular aggregation/activation functions (detailed in Sections~\ref{sec:back:gnn} and~\ref{sec:GNNs}). Moreover, we keep original GCN and g-SAGE as 2-layered models, while we adopt 5-layers for GIN (as proposed by the authors)\footnote{Obviously the number of layers could be increased/equalized for all of the models, however we keep them distinct to also accentuate their learning differences further.}. We further use the same latent dimension $d=10$ for all the weights in all the models. Finally we set average-pooling operation, followed by a single linear layer, as the final graph-level readout for prediction in each of the models.

\begin{figure*}[t!]
\centering
\resizebox{1.0\textwidth}{!}{
	\includegraphics[]{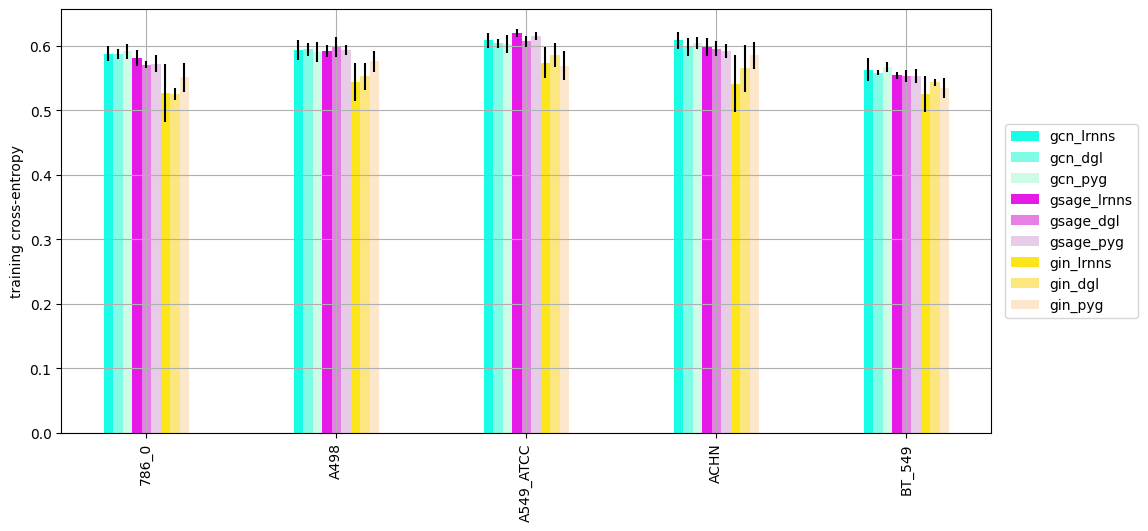}
}
\caption{Alignment of training errors of the 3 models (GCN, g-SAGE, GIN), as implemented in the 3 different frameworks (LRNNs, DGL, PyG), over 5 datasets.}
\label{fig:training}
\end{figure*}

While the declarative templating takes a very different approach from the procedural GNN frameworks, for the specific case of GNN templates it is easy to align their computations, as they are mostly simple sequential applications of the (i) neighborhood aggregation, (ii) weighting, and (iii) non-linear activation, which can be covered altogether by a single rule (Section~\ref{sec:GNNs})\footnote{However for a precise correspondence, care must be taken to respect the same order of the (i-iii) operations, which often varies across different reports and implementations.}.
\\\\
For the comparison, we choose the NCI~\citep{ncigi} molecular datasets\footnote{available at \url{ftp://ftp.ics.uci.edu/pub/baldig/learning/nci/gi50/}}, each containing thousands of molecules labeled by their ability to inhibit growth of different types of tumors. Note we only use the basic (Mol2~\citep{tripos2007tripos}) types of atoms and bonds without extra chemical features. For visual clarity we present results only for the first 5 of the total 73 datasets in alphabetical order (while we note that the results are very similar over the whole set). We use the same 10-fold crossvalidation splits for all the models. We further use Glorot initialization scheme~\citep{glorot2010understanding} where possible, and optimize using ADAM with a learning rate of $lr=1.5e^{-5}$ (betas and epsilon kept the usual defaults) against binary crossentropy over $2000$ epochae. Note that some other works propose a more radical training scheme with $lr=0.01$ and exponential decay by $0.5$ every 50 epochae~\citep{xu2018powerful}, however we find GNN training in this setting highly unstable\footnote{as is e.g. also visible in the respective Fig.4 reported in~\citep{xu2018powerful}.}, and thus unsuitable for the alignment of the different implementations. We then report the actual training errors (as opposed to accuracy) as the most consistent evaluation metric for the alignment purpose in Figure~\ref{fig:training}. While it is very difficult to align the training perfectly due to the underlying stochasticity, we can see that the performances are tightly aligned within a margin of standard deviation calculated over the folds. The differences are generally highest for the most complex GIN model, which also exhibits most unstable performance over the folds. Importantly, the differences between LRNNs and the other frameworks is generally not greater than between PyG and DGL themselves, which both utilize the exact same PyTorch modules and operations.

\subsection{Computing Performance}

The main aim of the declarative LRNN framework is quick prototyping of models aiming to integrate deep and relational learning capabilities, for which it generally provides more expressive constructs than that of GNNs (Section~\ref{sec:GNNs}) and does not contain any specific optimizations for computation over graph data. Additionally, it introduces a startup model compilation overhead as the particular models are not specified by the user but rather automatically induced by the theorem prover.
Moreover, it implements the neural training in a rather direct (but flexible) fashion of actual traversal over each network (such as in Dynet~\citep{neubig2017dynet}), and does so without batching, efficient tensor multiplication or GPU support\footnote{However it is possible to export the networks to be trained by any neural backend rather than the native Java engine.}. Nevertheless, we show that the increased expressiveness does not come at the cost of computation performance.

\begin{table}[]
\centering
\caption{Training times \textit{per epocha} across the different models and frameworks. Additionally, the startup model creation time (theorem proving) overhead of LRNNs is displayed.}
\begin{tabular}{c|c|c|c|c}
model/engine & LRNNs (s)       & PyG (s)        & DGL (s)        & LRNNs startup (s) \\
\hline
\hline
GCN          & \textbf{0.25 $\pm$ 0.01} & 3.24 $\pm$ 0.02  & 23.25 $\pm$ 1.94 & 35.2 $\pm$ 1.3        \\
\hline
g-SAGE        & \textbf{0.34 $\pm$ 0.01} & 3.83 $\pm$ 0.04  & 24.23 $\pm$ 3.80 & 35.4 $\pm$ 1.8        \\
\hline
GIN          & \textbf{1.41 $\pm$ 0.10} & 11.19 $\pm$ 0.06 & 52.04 $\pm$ 0.41 & 75.3 $\pm$ 3.2         
\end{tabular}
\label{tab:times}
\end{table}

\begin{figure*}[t!]
\centering
\resizebox{1.0\textwidth}{!}{
	\includegraphics[]{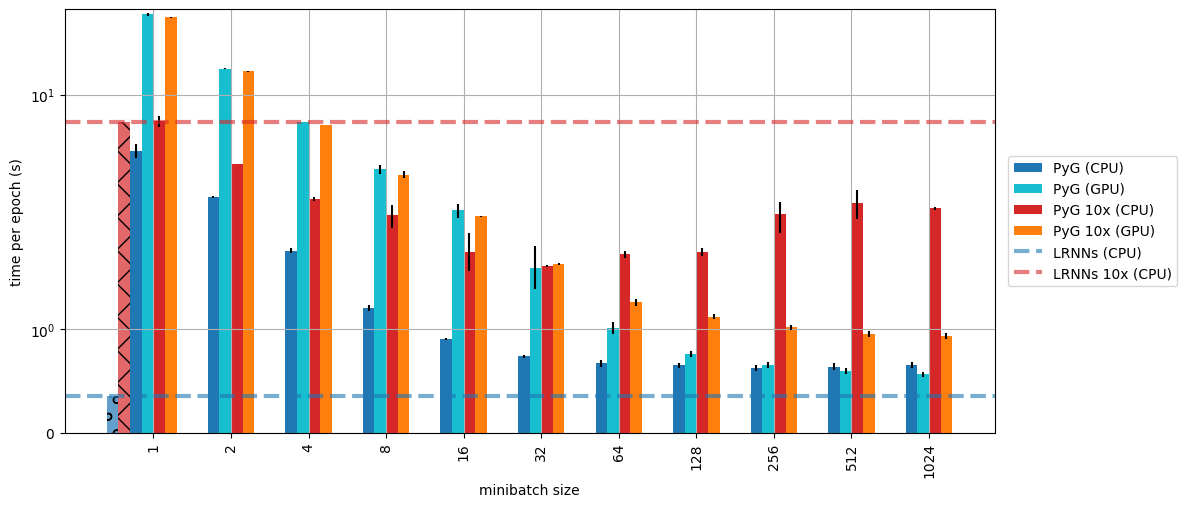}
}
\caption{Improving the computing performance of PyG via mini-batching and model size blow-up. The actual GCN model ($10\times10$ parameter matrices) and 10x inflated version ($100\times100$ parameter matrices) as run on CPU and GPU. Compared to a non-batched (batch=1) LRNN run on CPU.}
\label{fig:batching}
\end{figure*}

We evaluate the training times of a GCN over 10 folds of a single dataset (containing app. 3000 molecules) over the different models. We set Pytorch as the DGL backend (to match PyG), and train on CPU\footnote{We evaluated the training on CPU as in this problem setting the python frameworks run \textit{slower} on GPU (Figure~\ref{fig:batching}).} with a vanilla SGD (i.e. batch size=1) in all the frameworks. From the results in Table~\ref{tab:times}, we see that LRNNs surprisingly train significantly faster than PyG, which in turn runs significantly faster than DGL. While the performance edge of PyG over DGL generally agrees with~\citep{fey2019fast}\footnote{We note that we run both frameworks in default configurations, and there might be settings in DGL for which it does not lag behind PyG so rapidly. Note that for the small models of GCN and g-SAGE it is 10x slower, while for the bigger GIN model only 5x, which is in agreement with DGL's focus on large scale optimization.}, the (app. 10x) edge of LRNN seems unexpected, even accounting for the startup theorem proving overhead for the model creation (giving PyG a head start of app. 10 epochae).
We account the superior performance of LRNNs to the rather sparse, irregular, small, dynamic graphs for which the common vectorization techniques, repeatedly transforming the tensors there and back, often create more overhead than speedup, making it more efficient to traverse the actual spatial graph representations~\citep{neubig2017dynet}. Additionally, LRNNs are implemented in Java, removing the Python overhead, and contain some generic novel computation compression~\citep{vsourek2020lifting} techniques (providing about 3x speedup for the GNN templates).

Note we also prevented the frameworks from batching over several graphs, which they do by embedding the adjacency matrices into a block-diagonal matrix.
While (mini) batching has been shown to deteriorate model generalization~\citep{masters2018revisiting,wilson2003general}, it still remains the main source of speedup in deep learning frameworks~\citep{keskar2016large}, and is a highlighted feature of PyG, too. We show the additional PyG speedup gained by batching in Figure~\ref{fig:batching}. While batching truly boosts the PyG performance significantly, it still lacks behind the non-batched LRNNs\footnote{While LRNNs currently do not support batching natively, it can be emulated by outsourcing the training to Dynet.}. For illustration, we additionally include an inflated version of the GCN model by a 10x increase of all the tensor dimensionalities. In this setting we can finally observe a performance edge from mini-batching, due to vectorization and GPU\footnote{Also note that we used a non-high-end Ge-Force 940m, and the performance boost could thus be even more significant.}, over the non-batched LRNNs\footnote{On the other hand note that $dim=100$ is considerably high. Most implementations we found were in range \{8,16,32\} and we did not observe any \textit{test} performance improvement beyond $dim=5$ with the reported datasets and models.}.



\subsection{Model Generalization}

\begin{figure*}[t!]
\centering
\resizebox{1.0\textwidth}{9.5cm}{
	\includegraphics[]{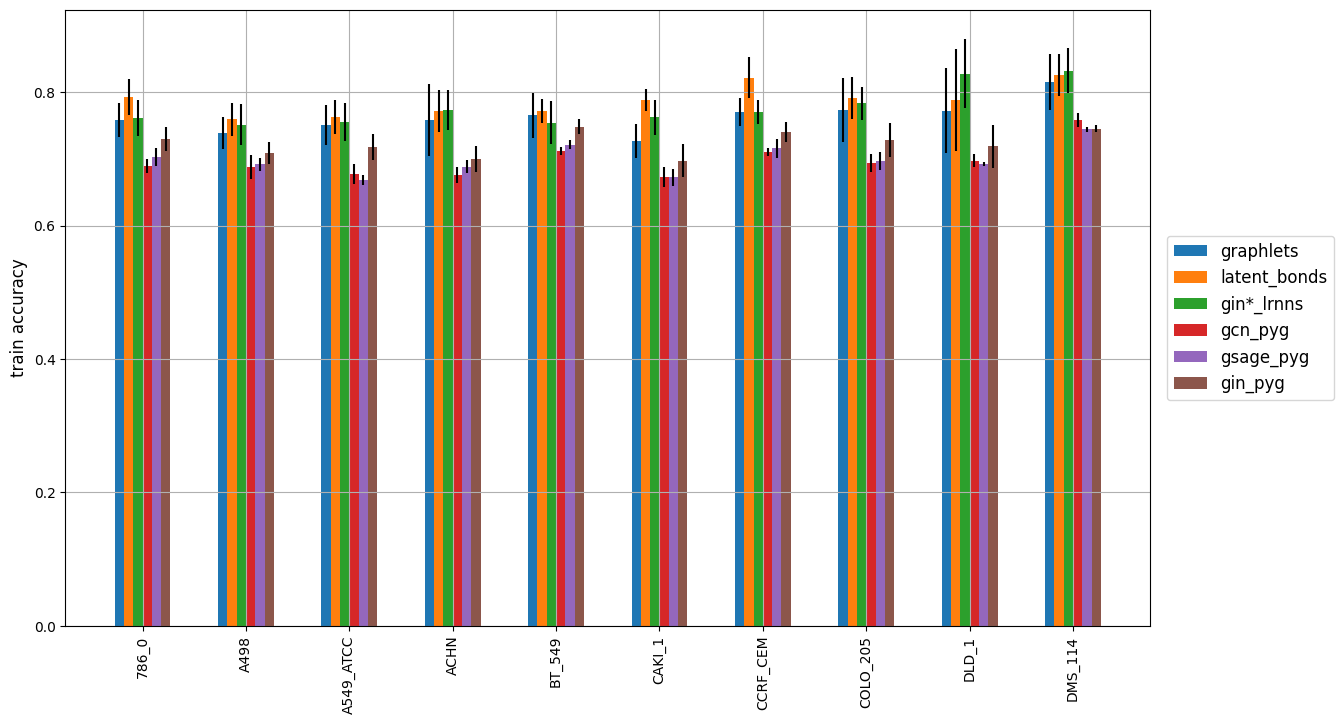}
}
\caption{Comparison of \textit{train} accuracies of selected models across 10 datasets.}
\label{fig:div_training}
\end{figure*}

\begin{figure*}[t!]
\centering
\resizebox{1.0\textwidth}{9.5cm}{
	\includegraphics[]{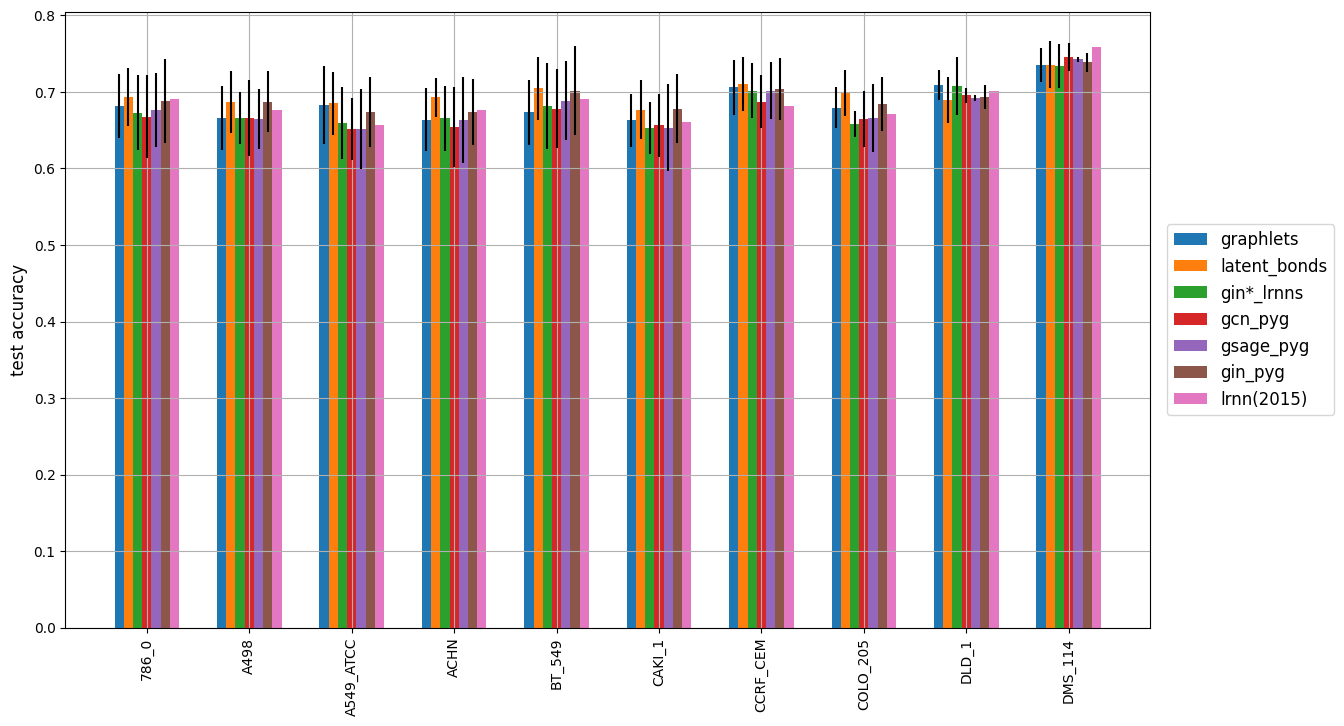}
}
\caption{Comparison of \textit{test} accuracies of selected models across 10 datasets.}
\label{fig:testing}
\end{figure*}

Finally we evaluate learning performances of the different models. We select the discussed GNN models of GCN, g-SAGE and GIN (we keep only the PyG implementation for clarity), and we further include some example relational templates for demonstration.
Particularly, we extend GIN with edge (bond) representations and associate all literals in all rules with learnable matrices (Sec.~\ref{sec:GNNs}), denoted as ``gin*''. In a second template we add a layer of graphlets (motifs) of size 3, aggregating jointly representations of \textit{three} neighboring nodes, on top of GIN, denoted as ``graphlets''. Lastly, we introduce latent bond learning (Sec.~\ref{sec:beyond}) into GIN, where bond (edge) representations are also aggregated into latent hierarchies, denoted as ``latent\_bonds''. Note that we restrict these new relational templates to the same tensor dimensionalities and number of layers as GIN. For statistical soundness, we increase the number of datasets to the first 10 (alphabetically). We run all the models on the same 10-fold crossvalidation splits with a 80:10:10 (train:val:test) ratio, and keep the same, previously reported, training hyperparameters. We display the aggregated \textit{training} errors in Figure~\ref{fig:div_training}, and the cross-validated \textit{test} errors, corresponding to the best validation errors in each split, in Figure~\ref{fig:testing}. 

We can observe that the training performance follows intuition about capacity of each model, where the more complex models generally dominate the simpler ones. However, the increased capacity does not seem to consistently translate to better test performance (contrary to the intuition stated in~\citep{xu2018powerful}). While we could certainly pick a subset of datasets to support the same hypothesis on test sets, we can generally see that none of the models actually performs consistently better than another, and even the simplest models (e.g. GCN) often outperform the ``powerful'' ones (GIN and its modifications), and the test performances are thus generally inconclusive\footnote{We note that the conclusion could be different for different types of datasets.}. While this is in contrast with the self-reported results accompanying the diverse GNN proposals on similar-sized graph datasets, it is in agreement with another (much larger) recent benchmark~\citep{dwivedi2020benchmarking}.

Additionally, we include results of an old LRNN template reported in~\citep{sourek2015lifted}, denoted as ``lrnn(2015)''. It was based on small graphlets of size 3, similarly to the ``graphlets'' template (and similarly to some other works~\citep{tu2019gl2vec,sankar2017motif}), however it only contained a single layer of learnable parameters for the atom and bond representations. Note that we use results from the original paper~\citep{sourek2015lifted} experiments, which were run with different hyperparameters and splits. Nevertheless, we can see that it is again generally on par with performance of the more recent, deeper, and bigger GNN models.



\section{Related Works}
\label{sec:related}

This work can be seen as a simple extension of the Lifted Relational Neural Networks~\citep{vsourek2015lifted} language by increasing the amount of (tensor) parameterization. In turn LRNNs were inspired by lifted graphical models~\citep{LIFTED} such as Bayesian Logic Programs~\citep{kersting2001bayesian} or Markov Logic Networks~\citep{richardson2006markov}, working in a probabilistic setting. From another view, LRNNs can also be seen as a direct extension of KBANN~\citep{KBANN} into relational setting. 
The most closely related works naturally comprise of other differentiable programming languages with relational expressiveness~\citep{de2020statistical}\footnote{Note that common differentiable programming frameworks, such as PyTorch or TensorFlow, are effectively propositional. They provide sets of evaluation functions (modules), with predefined hooks for backward differentiation, that can be assembled by users into differentiable programs in a procedural fashion. In contrast, with relational programming, such programs are firstly automatically assembled from the declarative template (by a theorem prover), and only then evaluated and differentiated in the same fashion. Such approach could thus be understood as ``meta-programming''~\citep{visser2002meta,hill1998meta} from the perspective of the current procedural frameworks.}.

There is a number of works targeting similar abilities by extending logic programming with numerical parameters. The most prominent framework in this category is the language of Problog \citep{de2007problog}, where the parameters and values further posses probabilistic interpretation. The extension of Deep-Problog~\citep{manhaeve2018deepproblog} then incorporates ``neural predicates'' into Problog programs. Since probabilistic logic programs can be commonly differentiated~\citep{fadja2017deep} and trained as such, the gradients can be passed from the logic program to the neural modules and trained jointly. While this is somewhat similar to LRNNs, Deep-Problog introduces a clear separation line between the neural and logical parts of the program, which communicate merely through the gradient values (and so any gradient-based learner could be used instead). The logical part with relational expressiveness is thus completely oblivious of structure of the gradient-ingesting learner and vice versa, and it is thus impossible to model complex convolutional patterns (i.e. relational patterns in the neural part) as demonstrated in this paper. On the other hand LRNNs do not have probabilistic interpretation. Related is also a recent extension of kProblog~\citep{orsini2017kproblog}, proposing integration of algebraic expressions into logic programs towards more general tensor-algebraic and ML algorithms.


Neural Theorem Provers (NTPs)~\citep{rocktaschel2017end} share very similar idea by the use of a relational logic template with a theorem prover to derive ground computation graphs, which are differentiable under certain semantics inspired by fuzzy logic. The use of parameterization differs between the frameworks, where NTPs are focused on learning embeddings of constants and LRNNs on embeddings of whole relational constructs\footnote{Note that this includes learning embeddings of constants, too, as demonstrated in some of the example templates.}. Nevertheless NTPs represent \textit{all} constants as embedding vectors, for which the theorem prover cannot perform standard unification, and NTPs thus resort to ``soft-unification'', effectively trying all possible constant combinations in the inference process. This prevents from using NTPs for explicit modelling of the exemplified convolutional neural architectures, and also severely limits NTP's scalability, where the latter has been partially addressed by some recent NTP extensions~\citep{Minervini2018,weber2019nlprolog}. LRNNs are more flexible in this sense as one can use the parameterization to specify which parts of the program keep the logical structure and which parts should succumb themselves to the exhaustive numerical optimization (and to combine them arbitrarily), enabling to find a more fine-grained neural-symbolic trade-off.

Another line of work is focused on inducing Datalog programs with the help of numerical relaxation. While such a task has traditionally been addressed by the means of Inductive Logic Programming (ILP)~\citep{muggleton1994inductive}, extending the rules with weights can help to relax the combinatorial search into a gradient descend optimization, while providing robustness to noise. An example of such an approach is $\delta$ILP~\citep{Evans2017}. Similarly to LRNNs, Datalog programs are unfolded by chaining the rules, where the associated parameters are trained against given target to be solved by the program. The parameterization in these approaches is used differently as its purpose is to determine the right \textit{structure} of the template. This is typically done by exhaustive enumeration from some restricted set of possible literal combinations (particularly 2 literals with arity at most 2 and no constants for $\delta$ILP), where each combination is then associated with a weight to determine its appropriateness for the program via gradient descend. The differentiability is again based on replacing the logical connectives with fuzzy logic operators (particularly product t-norm). Similarly, programs in a restricted subset of Datalog are learned in a system called TensorLog~\citep{cohen2016tensorlog,yang2017differentiable}, which is a differentiable probabilistic database based on belief propagation (limited to tree-like factor graphs). Another recently proposed related system is called Difflog~\citep{raghothaman2019difflog}, where the candidate rules are also exhaustively generated w.r.t. a more narrow language bias, thanks to which it seem to scale beyond previous systems. While we explicitly address only parameter learning in this paper, structure learning of the LRNN programs can also be done~\citep{vsourek2017stacked}.

Other class of approaches target full first order logic by providing mapping of all the logical constructs into numerical (tensor) spaces (``tensorization''~\citep{garcez2019neural}). For instance, one can cast constants to vectors, functions terms to vector functions of the corresponding dimensionality, and similarly predicates to tensors of the corresponding arity-dimension~\citep{Rocktaschel2015a,diligenti2017semantic}. Again adopting a fuzzy logic interpretation of the logical connectives, the learning problem can then be cast as a constrained numerical optimization problem, including works such as Logic Tensor Networks~\citep{serafini2016logic} or LYRICS~\citep{marra2019lyrics}. While the distributed representation of the logical constructs is the subject of learning, in contrast with the discussed Datalog program structure learning approaches, the weight (strength) of each rule needs to be specified apriori -- a limitation which was recently addressed in~\citep{marra2020relational}. Other recent works based on the idea of fully dissolving the logic into tensors, moving even further from the logical interpretation, include e.g. Neural Logic Machines~\citep{dong2019neural}. While these frameworks are theoretically more expressive than LRNNs (lacking the function terms and non-definite clauses), the whole logic interpretation is only approximate and completely dissolved in the tensor weights in these frameworks. Consequently, they lack the capability of precise relational logic inference chaining which we use to explicitly model the advanced convolutional neural structures, such as GNNs, in this paper.

Alternatively, LRNNs can be seen as an extension of GNNs, as discussed in this paper. From the graph-level perspective, the most similar idea to the introduced relational templating has become popular in the knowledge discovery community as \textit{``meta-paths''}~\citep{dong2017metapath2vec,huang2017heterogeneous} defined on the schema-level of a heterogeneous information network. A meta-path is simply a sequence of types, the concrete instantiations of which are then searched for in the ground graphs. Such ground sequences can then be used to define node similarities~\citep{sun2011pathsim,shang2016meta}, random walks~\citep{dong2017metapath2vec} as well as node embeddings~\citep{shi2018heterogeneous,fu2017hin2vec}. An extension from paths to small DAGs was then proposed as ``meta-graph'' (or ``meta-structure'')~\citep{huang2016meta,sun2018joint}. Any meta-path or meta-graph can be understood as conjunctive a rule in a LRNN template. Naturally, we can stack multiple meta-graphs to create deep hierarchies and, importantly, differentiate them through to jointly learn all the parameters, and provide further extensions towards relational expressiveness exemplified in this paper.


\section{Conclusions}
\label{sec:concl}

We introduced a differentiable declarative programming approach for specification of advanced relational neural architectures, based on the language of Lifted Relational Neural Networks (LRNNs) \citep{sourek2018lifted}. We demonstrated how simple parameterized logic programs, also called templates, can be efficiently used for declaration and learning of complex convolutional models, with a particular focus on Graph Neural Networks~(GNNs). In contrast with the commonly used procedural (Python) frameworks, LRNNs abstract away the creation of the specific computation graphs, which are dynamically unfolded from the template by an underlying theorem prover. As a result, creating a diverse class of complex neural architectures reduces to rather trivial modifications of the templates, distilling only the high level idea of each architecture. We illustrated versatility of the approach on a number of examples, ranging gradually from simple neural models to complex GNNs, including very recent GNN models and their extensions. Finally we showed how the existing models can be easily extended to even higher relational expressiveness.

In the experiments, we then demonstrated correctness and computation efficiency by the means of comparison against modern deep learning frameworks. We showed that while LRNNs are designed with main focus on flexibility and abstraction, they do not suffer from computation inefficiencies for the simpler (GNN) models, as one might expect. On the contrary, we demonstrated that for a range of existing GNN models and their practical parameterizations, LRNNs actually outperform the existing frameworks optimized specifically for GNNs.

While there is a number of related works targeting the integration of deep and relational learning, to our best knowledge, capturing advanced convolutional neural architectures in an exact manner, as exemplified in this paper, would not be possible with the other approaches. The proposed relational upgrades can then be understood as proper extensions of the existing, arguably popular, GNN models. However, we showed that generalization performance of various state-of-the-art GNN models is somewhat peculiar, as they actually performed with rather insignificant test error improvements, when measured uniformly over a large set of medium-sized, molecular structure-property prediction datasets, which is in agreement with another recent benchmark~\citep{dwivedi2020benchmarking}.


\newpage

\appendix

\section{Appendix}

\subsection{Differences from LRNNs
~\citep{sourek2018lifted}
}
\label{sec:appendix}

\begin{algorithm}[h]
\caption{Transforming ground neural network into vectorized form}
	\begin{algorithmic}[1]
	\Function{vectorize}{neurons}
	    \State $\mathcal{N} \gets \bigcup neurons$
		\State $(depth,\mathcal{N}) \gets topologicOrder(\mathcal{N}$)
		\State $Layers = \varnothing$
		\For{$i = 1 : depth$}
		\State $ M_i \gets initMatrix() $
		\State $\mathcal{M} \gets neuronsAtLevel(i,\mathcal{N})$
    		\For{$neuron \in \mathcal{M}$}
    		    \State $(inputs, weights) \gets inputs(neuron)$
    		    \For{$(input, weight) \in (inputs, weights)$}
        		    \If{$ getLevel(input) = i+1 $}
        		        \State $M_i(neuron,input) = weight $
        		    \Else 
        		        \State $skipConnect \gets void(neuron, i+1)$
        		        \State $M_i(neuron,skipConnect) = 1 $
        		    \EndIf
    		    \EndFor
    		\EndFor
    		\State $Layers = Layers \cup M_i$
		\EndFor
		\State \Return $Layers$
	\EndFunction
	\end{algorithmic}
	\label{alg:vectorization}
\end{algorithm}

{T}he {framework}
introduced {in this paper} closely follows the original LRNNs~\citep{sourek2018lifted}. In fact, the main semantic difference is {``}merely{''} in the parameterization of the rules, where one can now include the weights within the bodies (conjunctions), too, e.g.
$$w^{(2)}_1 :: {node^{(2)}_1(X)} \leftarrow w^{(1)}_1 \cdot node^{(0)}_1(X) \wedge w^{(1)}_2 \cdot node^{(0)}_2(X)\\$$
We note that this could be in essence emulated in the original LRNNs through the use of auxiliary predicates, such as in~\citep{sourek2018lifted}, as follows
\begin{align*}
w^{(2)}_1 &: {neuron^{(2)}_1(X)} \leftarrow neuron^{(1)}_1(X) \wedge neuron^{(1)}_2(X)&\\
w^{(1)}_1 &: {neuron^{(1)}_1(X)} \leftarrow neuron^{(0)}_1(X)&\\
w^{(1)}_2 &: {neuron^{(1)}_2(X)} \leftarrow neuron^{(0)}_2(X)&
\end{align*}
which might be more appropriate in scenarios where the neurons correspond to actual logical concepts under fuzzy logic semantics\footnote{Note that any model from the original LRNNs can still be directly encoded in the new formalism.}, while the second representation is arguably more suitable to exploit the correspondence with standard deep learning architectures.

{Another difference is that} we {now also} allow tensor weights and values. While these {could} {be} modeled in LRNNs, too, for instance in the ``soft-clustering'' (embedding) construct~\citep{vsourek2015lifted} used for atom representation learning:
\begin{align*}
w_{o_1} &: \textit{gr}_1(X) \leftarrow {O}(X)&
& ... &
w_{h_1} &: \textit{gr}_1(X) \leftarrow {H}(X)\\
w_{o_2} &: \textit{gr}_2(X) \leftarrow {O}(X)&
& ... &
w_{h_2} &: \textit{gr}_2(X) \leftarrow {H}(X)
\end{align*}
the {tensor}-valued weights offer an arguably more elegant representation of the same construct:
\begin{align*}
[w_{o_1},w_{o_2}]::\textit{gr}(X) \leftarrow {O}(X) &~~~~~~~~~~~...~~~~& [w_{h_1},w_{h_2}]::\textit{gr}(X) \leftarrow {H}(X)\\
\end{align*}
In general, with the new representation we can merge scalar weights of individual neurons into tensors used by the \textit{nodes} (prev. referred to as ``neurons'') in the computation graph. Note that we can process any ground LRNN network {into this form}, i.e. turn individual neurons into matrix layers, in a similar manner, as outlined in Algorithm~\ref{alg:vectorization}.

Being heavily utilized in deep learning, such transformation can significantly speed up the training of the networks. However by reducing the number of rules, effectively merging together semantically equivalent rules which do not differ up to their (scalar) parameterization, we can also alleviate much of the complexity during model creation, i.e. calculation of the least Herbrand model, by avoiding repeated calculations. This results in a significant speedup during the model creation process.

\subsubsection{Network Pruning}
\label{sec:app:pruning}
\begin{algorithm}[t]
\caption{Pruning linear chains of unnecessary transformations}
	\begin{algorithmic}[1]
	\Function{prune}{neurons}
	    \State $\mathcal{N} \gets \bigcup neurons$
	    \For{$neuron \in \mathcal{N}$}
	        \State $(inputs, weights) \gets inputs(neuron)$
	        \If{$inputs.size = 1 \wedge weights = \varnothing$}
	            \State $outputs \gets outputs(neuron)$
	            \For{$output \in outputs$}
	                \State $input = inputs[0]$
	                \State $output.replaceInput(neuron,input)$
	                \State $input.replaceOutput(neuron,output)$
	            \EndFor
	        \EndIf
	    \EndFor
	    \State \Return $connectedComponent(\mathcal{N})$
    \EndFunction
	\end{algorithmic}
	\label{alg:pruning}
\end{algorithm}
Following the computation graph creation procedure from Section~\ref{sec:semantics}, we might end up with unnecessary trivial neural transformations through auxiliary predicates in cases, where the original rules have only a single literal in body and are unweighted. For mitigation, we can apply a straightforward procedure for detection and removal of linear chains of these trivial operations, as described in Algorithm~\ref{alg:pruning}. While such an operation arguably changes the inference and logical semantics of the original model, these structures do not contribute to learning capacity of the model and, on the contrary, cause gradient diminishing. This technique is thus particularly suited for improving training performance in correspondence with standard deep learning architectures. While this form of pruning can be theoretically performed directly in the template, it is easier to do as a post-processing step in the resulting neural networks.
\\\\
Finally, the new LRNN framework (``NeuraLogic'') presents a completely new \textit{implementation}\footnote{available at \url{https://github.com/GustikS/NeuraLogic}} of the idea, with the whole functionality build from scratch, while aiming at flexibility and modularity.

\newpage


%
%

\bibliographystyle{spbasic}      

\bibliography{references}

\end{document}

%% file: img/settings.tex
\tikzstyle{atom}  =  [circle, scale=1.7, circle, shading=ball]

\tikzset{my label/.style args={#1:#2}{
  append after command={
    (\tikzlastnode.center) node [#1] {#2}
    }
  }
}

\tikzstyle{edgenode}  =  [thin, draw=black, align=center,fill=white,font=\small]
\tikzstyle{edgeweight}  =  [near start, above, sloped,outer sep=3pt, inner sep=1pt ,fill=white,font=\large]
\tikzstyle{gredge}  =  [outer sep=6pt, inner sep=1pt ,fill=white]
\tikzstyle{gredge1}  =  [outer sep=2pt, inner sep=1pt ,fill=white]

\tikzstyle{kappa} = [
  draw, scale = 1.5,  minimum size=1cm, circle, rounded corners,shading=radial,inner color=white,
  minimum height=1cm,
  align=center
  ]
  
\tikzstyle{lambda} = [
  draw, scale = 1.5,  minimum size=1cm, circle, rounded corners,shading=radial,inner color=white,
  minimum height=1cm,
  align=center
  ]

%% file: img/tikz/template.tex
\begin{tikzpicture}
[node distance=2.5cm and 0.5cm,
mynode/.style={
  draw, scale = 1.4,  minimum size=1cm, circle, rounded corners,shading=radial,outer color=gray!30,inner color=white,
  minimum height=1cm,
  align=center
  },
myfact/.style={
  draw, scale = 1.7,  minimum size=1cm, rounded corners,left color=white,
  minimum height=1cm,
  align=center
  }
]

\begin{scope}[xshift=-3.3cm]

\node[scale=1.3] (hydro) at (-5.4,-1.2){\input{img/water}};

\node[scale=1.5] (hydro) at (-5.2,3){\large sample 1:};
\node[scale=1.5] (hydro) at (-5,7){\large template:};

\begin{scope}[xshift=8cm, yshift=7cm]
\node[myfact, color=white, scale = 1.5] (bond) {\textcolor{red}{$\alpha_1$ :} \textcolor{violet}{\textbf{W$_{h1}$}} \textcolor{red}{::} h(X) $\impl$ \textcolor{magenta}{\textbf{W$_a$}} \textcolor{red}{:} a(Y) $\conj$ \textcolor{orange}{\textbf{W$_b$}} \textcolor{red}{:} b(X,Y) \textcolor{red}{.}};

\node[myfact, color=white, scale = 1.5] (embed2) [right=1cm of bond]{\textcolor{blue}{$\alpha_2$ :} \textcolor{teal}{\textbf{W$_q$}} \textcolor{red}{::} q $\impl$ \textcolor{cyan}{\textbf{W$_{h2}$}} \textcolor{red}{:} h(X) \textcolor{red}{.}};
\end{scope}

\node [draw=red!30, inner sep=5pt, fit={(bond)}] (rec1) {};
\node [draw=blue!30, inner sep=5pt, fit={(embed2)}] (rec2) {};


\end{scope}

\begin{scope}[yshift=1.3cm, xshift=0cm]

\begin{scope}[xshift=0cm, yshift=2cm, node distance=1.3cm and 0cm]
\node[myfact, label={[xshift=-1.6cm, yshift=0.4cm]{\LARGE Fact nodes}}, right color=brown!40!white] (f1) {$b(h_1,o_1)$};
\node[myfact, right color=brown!40!white] (f7) [left=0.9cm of f1] {$b(o_1,h_1)$};
\node[myfact, right color=black!40!white] (f2) [below left=0.5 and -0.4cm of f1] {$a(h_1)$};
\node[myfact, right color=black!40!white] (f3) [below of=f2] {$a(o_1)$};
\node[myfact, right color=black!40!white] (f4) [below of=f3] {$a(h_2)$};
\node[myfact, right color=brown!40!white] (f5) [below right=0.5 and -0.4cm of f4] {$b(h_2,o_1)$};
\node[myfact, right color=brown!40!white] (f6) [left=0.9cm of f5] {$b(o_1,h_2)$};

\end{scope}

\begin{scope}[xshift=6cm, yshift=1cm, node distance=1.3cm and 0cm]
\node[myfact, label={[xshift=0cm, yshift=0.4cm]{\LARGE Rule nodes}}, right color=red!20!white, my label={font = \Large, black,below=0.3cm:$\wedge$}] (r1) {R$_{\alpha{_1}\theta_1}^{h(h_1)}[X/h_1,Y/o_1]$};
\node[myfact, right color=red!20!white, my label={font = \Large, black,below=0.3cm:$\wedge$}] (r2) [below of=r1] {R$_{\alpha{_1}\theta_2}^{h(o_1)}[X/o_1,Y/h_1]$};
\node[myfact, right color=red!20!white, my label={font = \Large, black,below=0.3cm:$\wedge$}] (r3) [below of=r2] {R$_{\alpha{_1}\theta_3}^{h(o_1)}[X/o_1,Y/h_2]$};
\node[myfact, right color=red!20!white, my label={font = \Large, black,below=0.3cm:$\wedge$}] (r4) [below of=r3]{R$_{\alpha{_1}\theta_4}^{h(h_2)}[X/h_2,Y/o_1]$};
\end{scope}

\begin{scope}[xshift=12cm, yshift=0cm, node distance=1.3cm and 0cm]
\node[myfact, label={[xshift=0.1cm, yshift=0.4cm]{\LARGE {Aggregation nodes}}},  right color=red!30!white, my label={font = \LARGE, black,below=0.3cm:$*$}] (g1) {G$_{\alpha_1}^{h(h_1)}[X/h_1]$};
\node[myfact, right color=red!30!white, my label={font = \LARGE, black,below=0.3cm:$*$}] (g2) [below of=g1] {G$_{\alpha_1}^{h(o_1)}[X/o_1]$};
\node[myfact, right color=red!30!white, my label={font = \LARGE, black,below=0.3cm:$*$}] (g3) [below of=g2] {G$_{\alpha_1}^{h(h_2)}[X/h_2]$};
\end{scope}

\begin{scope}[xshift=16.5cm, yshift=0cm, node distance=1.3cm and 0cm]
\node[myfact, label={[xshift=0.1cm, yshift=0.5cm]{\LARGE {Atom nodes}}},  right color=red!40!white, my label={font = \LARGE, black,below=0.3cm:$\vee$}] (a1) {A$_{h(h1)}$};
\node[myfact,  right color=red!40!white, my label={font = \LARGE, black,below=0.3cm:$\vee$}] (a2) [below of=a1] {A$_{h(o1)}$};
\node[myfact,  right color=red!40!white, my label={font = \LARGE, black,below=0.3cm:$\vee$}] (a3) [below of=a2] {A$_{h(h2)}$};
\end{scope}

\begin{scope}[xshift=21cm, yshift=0cm, node distance=1.3cm and 0cm]
\node[myfact, label={[xshift=0.1cm, yshift=0.4cm]{\LARGE {Rule nodes}}},  right color=blue!20!white, my label={font = \LARGE, black,below=0.3cm:$\wedge$}] (rr1) {R$_{\alpha{_2}\theta_5}^{q}[X/h_1]$};
\node[myfact,  right color=blue!20!white, my label={font = \LARGE, black,below=0.3cm:$\wedge$}] (rr2) [below of=rr1] {R$_{\alpha{_2}\theta_6}^{q}[X/o_1]$};
\node[myfact,  right color=blue!20!white, my label={font = \LARGE, black,below=0.3cm:$\wedge$}] (rr3) [below of=rr2] {R$_{\alpha{_2}\theta_7}^{q}[X/h_2]$};
\end{scope}

\begin{scope}[xshift=25.8cm, yshift=-2cm, node distance=1.3cm and 0cm]
\node[myfact, label={[xshift=0.1cm, yshift=0.4cm]{\LARGE {Aggregation nodes}}},  right color=blue!30!white, my label={font = \LARGE, black,below=0.3cm:$*$}] (gg1) {G$_{\alpha_2}^{q}[\varnothing]$};
\end{scope}

\begin{scope}[xshift=30cm, yshift=-2cm, node distance=1.3cm and 0cm]
\node[myfact, label={[xshift=0.1cm, yshift=0.4cm]{\LARGE {Atom nodes}}},  right color=blue!40!white, my label={font = \LARGE, black,below=0.3cm:$\vee$}] (q1) {A$_q$};
\end{scope}

\node [draw=red!30, inner sep=12pt, fit={(f7) (f6) (a1) (a3)}] (rec11) {};
\node [draw=blue!30, inner sep=12pt, fit={(a1) (a3) (q1)}] (rec21) {};


\draw [dashed,red!30,->] (rec1) edge node[gredge, above] {} (rec11);
\draw [dashed,blue!30,->] (rec2) edge node[gredge, above] {} (rec21);

\draw [orange,->] (f1) edge node[gredge, below] {\textcolor{orange}{$W_b$}} (r1);
\draw [magenta,->] (f3) edge node[gredge, below] {$W_a$} (r1);

\draw [orange,->] (f7) to[out=-25,in=160] node[gredge, below] {\textcolor{orange}{$W_b$}} (r2);
\draw [magenta,->] (f2) edge node[gredge, below] {$W_a$} (r2);

\draw [orange,->] (f6) to[out=25,in=-160] node[gredge, below] {\textcolor{orange}{$W_b$}} (r3);
\draw [magenta,->] (f4) edge node[gredge, below] {$W_a$} (r3);

\draw [orange,->] (f5) edge node[gredge, below] {\textcolor{orange}{$W_b$}} (r4);
\draw [magenta,->] (f3) edge node[gredge, below] {$W_a$} (r4);

\draw [gray,->] (r1) edge node[gredge, below] {} (g1);
\draw [gray,->] (r2) edge node[gredge, below] {} (g2);
\draw [gray,->] (r3) edge node[gredge, below] {} (g2);
\draw [gray,->] (r4) edge node[gredge, below] {} (g3);

\draw [violet,->] (g1) edge node[gredge, below] {$W_{h1}$} (a1);
\draw [violet,->] (g2) edge node[gredge, below] {$W_{h1}$} (a2);
\draw [violet,->] (g3) edge node[gredge, below] {$W_{h1}$} (a3);

\draw [cyan,->] (a1) edge node[gredge, below] {$W_{h2}$} (rr1);
\draw [cyan,->] (a2) edge node[gredge, below] {$W_{h2}$} (rr2);
\draw [cyan,->] (a3) edge node[gredge, below] {$W_{h2}$} (rr3);

\draw [gray,->] (rr1) edge node[gredge, below] {} (gg1);
\draw [gray,->] (rr2) edge node[gredge, below] {} (gg1);
\draw [gray,->] (rr3) edge node[gredge, below] {} (gg1);

\draw [teal,->] (gg1) edge node[gredge, below] {$W_q$} (q1);

\end{scope}


\node[scale=1.3] (hydro) at (-8.4,-12){\input{img/hydrogen}};

\node[scale=1.5] (hydro) at (-8.4,-8.8){\large sample 2:};

\begin{scope}[yshift=-11cm, xshift=0cm]

\begin{scope}[xshift=-2cm, yshift=2.2cm, node distance=1.3cm and 0cm]
\node[myfact, label={[xshift=0cm, yshift=0.4cm]{\LARGE Fact nodes}}, right color=brown!40!white] (f1) {$b(h_1,h_2)$};
\node[myfact, right color=black!40!white] (f2) [below  of= f1] {$a(h_2)$};
\node[myfact, right color=black!40!white] (f3) [below of=f2] {$a(h_1)$};
\node[myfact, right color=brown!40!white] (f4) [below  of= f3] {$b(h_2,h_1)$};

\end{scope}

\begin{scope}[xshift=6cm, yshift=0cm, node distance=1.3cm and 0cm]
\node[myfact, label={[xshift=0cm, yshift=0.4cm]{\LARGE Rule nodes}}, right color=red!20!white, my label={font = \Large, black,below=0.3cm:$\wedge$}] (r1) {R$_{\alpha{_1}\theta_1}^{h(h_1)}[X/h_1,Y/h_2]$};
\node[myfact, right color=red!20!white, my label={font = \Large, black,below=0.3cm:$\wedge$}] (r2) [below of=r1]{R$_{\alpha{_1}\theta_2}^{h(h_2)}[X/h_2,Y/h_1]$};
\end{scope}

\begin{scope}[xshift=12cm, yshift=0cm, node distance=1.3cm and 0cm]

\node[myfact, label={[xshift=0.1cm, yshift=0.4cm]{\LARGE {Aggregation nodes}}},  right color=red!30!white, my label={font = \LARGE, black,below=0.3cm:$*$}] (g1) {G$_{\alpha_1}^{h(h_1)}[X/h_1]$};
\node[myfact, right color=red!30!white, my label={font = \LARGE, black,below=0.3cm:$*$}] (g2) [below of=g1] {G$_{\alpha_1}^{h(h_2)}[X/h_2]$};
\end{scope}

\begin{scope}[xshift=16.5cm, yshift=0cm, node distance=1.3cm and 0cm]
\node[myfact, label={[xshift=0.1cm, yshift=0.5cm]{\LARGE {Atom nodes}}},  right color=red!40!white, my label={font = \LARGE, black,below=0.3cm:$\vee$}] (a1) {A$_{h(h_1)}$};
\node[myfact,  right color=red!40!white, my label={font = \LARGE, black,below=0.3cm:$\vee$}] (a2) [below of=a1] {A$_{h(h_2)}$};
\end{scope}

\begin{scope}[xshift=21cm, yshift=0cm, node distance=1.3cm and 0cm]
\node[myfact, label={[xshift=0.1cm, yshift=0.4cm]{\LARGE {Rule nodes}}},  right color=blue!20!white, my label={font = \LARGE, black,below=0.3cm:$\wedge$}] (rr1) {R$_{\alpha{_2}\theta_3}^{q}[X/h_1]$};
\node[myfact,  right color=blue!20!white, my label={font = \LARGE, black,below=0.3cm:$\wedge$}] (rr2) [below of=rr1] {R$_{\alpha{_2}\theta_4}^{q}[X/h_2]$};
\end{scope}

\begin{scope}[xshift=25.8cm, yshift=-1cm, node distance=1.3cm and 0cm]
\node[myfact, label={[xshift=0.1cm, yshift=0.4cm]{\LARGE {Aggregation nodes}}},  right color=blue!30!white, my label={font = \LARGE, black,below=0.3cm:$*$}] (gg1) {G$_{\alpha_2}^{q}[\varnothing]$};
\end{scope}

\begin{scope}[xshift=30cm, yshift=-1cm, node distance=1.3cm and 0cm]
\node[myfact, label={[xshift=0.1cm, yshift=0.5cm]{\LARGE {Atom nodes}}},  right color=blue!40!white, my label={font = \LARGE, black,below=0.3cm:$\vee$}] (q2) {A$_q$};
\end{scope}

\node [draw=red!30, inner sep=12pt, fit={(f1) (f4) (a1) (a2)}] (rc11) {};
\node [draw=blue!30, inner sep=12pt, fit={(a1) (a2) (q2)}] (rc21) {};


\draw [dashed,red!30,->] (rec11) edge node[gredge, above] {} (rc11);
\draw [dashed,blue!30,->] (rec21) edge node[gredge, above] {} (rc21);

\draw [orange,->] (f1) edge node[gredge, below] {{$W_b$}} (r1);
\draw [magenta,->] (f2) edge node[gredge, below] {$W_a$} (r1);

\draw [orange,->] (f4) edge node[gredge, below] {{$W_b$}} (r2);
\draw [magenta,->] (f3) edge node[gredge, below] {$W_a$} (r2);

\draw [gray,->] (r1) edge node[gredge, below] {} (g1);
\draw [gray,->] (r2) edge node[gredge, below] {} (g2);

\draw [violet,->] (g1) edge node[gredge, below] {$W_{h1}$} (a1);
\draw [violet,->] (g2) edge node[gredge, below] {$W_{h1}$} (a2);

\draw [cyan,->] (a1) edge node[gredge, below] {$W_{h2}$} (rr1);
\draw [cyan,->] (a2) edge node[gredge, below] {$W_{h2}$} (rr2);

\draw [gray,->] (rr1) edge node[gredge, below] {} (gg1);
\draw [gray,->] (rr2) edge node[gredge, below] {} (gg1);

\draw [teal,->] (gg1) edge node[gredge, below] {$W_q$} (q2);

\end{scope}

\end{tikzpicture}

%% file: img/water.tex
\begin{tikzpicture}

\node[atom, ball color=red!50!white, label={[xshift=-0.7cm, yshift=-2.9cm]$b(o_1,h_2)$}, label={[xshift=-0.7cm, yshift=0cm]$b(o_1,h_1)$}, label=left:{$a(o_1)$}] at (0,1) (o1) {\huge \textbf{O\textsuperscript{{\tiny 1}}}};

\node[atom, circle, ball color=blue!50!white, label=above right:{$a(h_2)$}, label=below:{$b(h_2,o_1)$}] at (1,-0.1) (h1) {\textbf{H\textsuperscript{\tiny 2}}};
\node[atom, circle, ball color=blue!50!white, label=below right:{$a(h_1)$}, label=above:{$b(h_1,o_1)$}] at (1,2.1) (h2) {\textbf{H\textsuperscript{\tiny 1}}};

\end{tikzpicture}

%% file: img/tikz/MLPs.tex
\begin{tikzpicture}
[transform shape,rotate=0, node distance=2.0cm and 2.0cm,
ar/.style={->,>=latex},
mynode/.style={
  draw, scale = 1.0,  minimum size=1cm, rounded corners,left color=white,
  minimum height=1cm,
  align=center
  }
]

\tikzstyle{neuron}  =  [rectangle, draw, scale = 1.0,  minimum size=1cm]
\tikzstyle{treenode}  =  [mynode, right color=black!30!white]
\tikzstyle{recnode}  =  [mynode, right color=magenta!30!white]
\tikzstyle{qnode}  =  [mynode, right color=blue!30!white]
\tikzstyle{grbond}  =  [mynode, right color=black!30!white]
\tikzstyle{gratom}  =  [mynode]
\tikzstyle{grgroup} =  [mynode, right color=brown!30!white]
\tikzstyle{grexpl}  =  [mynode, right color=violet!30!white]
\tikzstyle{edgenode}  =  [thin, draw=black, align=center,fill=white,font=\small]

\node[treenode] (f1) {F$_{f,v}$};
\node[recnode,right color=red!30!white] (r1) [right of=f1] {R$_{\alpha_1\theta}^{h1}$};
\node[recnode,right color=red!30!white] (g1) [right of=r1] {G$_{\alpha_1}^{h1}$};
\node[recnode,right color=red!30!white] (a1) [right of=g1] {A$_{h1}$};

\node[recnode,right color=blue!30!white] (r2) [right of=a1] {R$_{\alpha_2\theta}^{h2}$};
\node[recnode,right color=blue!30!white] (g2) [right of=r2] {G$_{\alpha_2}^{h2}$};
\node[qnode,right color=blue!30!white] (a2) [right of=g2] {A$_{h_2}$};


\draw[gray,->] (f1) -> node[gray, below] {} (r1);
\draw[gray,->] (r1) -> node[gray, below] {} (g1);
\draw[orange,->] (g1) -> node[ below] {$\mathbf{W_1}$} (a1);

\draw[gray,->] (a1) -> node[gray, below] {} (r2);
\draw[gray,->] (r2) -> node[gray, below] {} (g2);
\draw[violet,->] (g2) -> node[ below] {$\mathbf{W_2}$} (a2);

\node[mynode, draw=white, label={[xshift=0.0cm, yshift=-1.1cm]{pruning}}] (prune) [right of=a2] {\Large $\rightarrow$};

\node[treenode] (ff1) [right of=prune] {F$_{f,v}$};
\node[recnode,right color=red!30!white] (aa1) [right of=ff1] {A$_{h1}$};
\node[qnode] (aa2) [right of=aa1] {A$_{h_2}$};

\draw[orange,->] (ff1) -> node[below] {$\mathbf{W_1}$} (aa1);
\draw[violet,->] (aa1) -> node[below] {$\mathbf{W_2}$} (aa2);

\end{tikzpicture}

%% file: img/network/CNN.tex
\begin{tikzpicture}
[transform shape,rotate=0, node distance=2.0cm and 2.0cm,
ar/.style={->,>=latex},
mynode/.style={
  draw, scale = 1.0,  minimum size=1cm, rounded corners,left color=white,
  minimum height=1cm,
  align=center
  }
]

\tikzstyle{neuron}  =  [circle, draw, scale = 1.0,  minimum size=1cm]
\tikzstyle{grbond}  =  [mynode, right color=red!30!white]
\tikzstyle{gratom}  =  [mynode]
\tikzstyle{grgroup} =  [mynode, right color=brown!30!white]
\tikzstyle{grexpl}  =  [mynode, right color=red!30!white]
\tikzstyle{edgenode}  =  [thin, draw=black, align=center,fill=white,font=\small]

\begin{scope}[xshift=0cm,yshift=0cm]

\begin{scope}[xshift=-1cm, yshift=-2cm]

\draw[step=2.0,black,thin] (0,0) grid (10,1.1);

\node at (1,0.5) {\Large $v_1$};
\node at (3,0.5) {\Large $v_2$};
\node at (5,0.5) {\Large $v_3$};
\node at (7,0.5) {\Large $v_4$};
\node at (9,0.5) {\Large $v_5$};

\node[] (gh2) [xshift=5cm, yshift=-0.7cm] {\LARGE image $I$ as a vector of pixel values};

\end{scope}

\begin{scope}[xshift=0cm]

\node[neuron,label={[xshift=-1.7cm, yshift=-0.5cm]{\large Features}},label={[xshift=-1.7cm, yshift=-1.2cm]{\large (pixels)}}] (gh1) {$f^1$};
\node[neuron] (gh2) [right of=gh1] {$f^2$};
\node[neuron] (gh3) [right of=gh2] {$f^3$};
\node[neuron] (gh4) [right of=gh3] {$f^4$};
\node[neuron] (gh5) [right of=gh4] {$f^5$};

\end{scope}

\begin{scope}[xshift=1cm, yshift=2cm]

\node[neuron,label={[xshift=-2.2cm, yshift=-0.6cm]{\large Filter-map}},label={[xshift=-2.2cm, yshift=-1.2cm]{\large (convolution)}}] (gr1h2) [above of=gh2] {$h^1$};
\node[neuron] (gr1h3) [right of=gr1h2] {$h^2$};
\node[neuron] (gr1h4) [right of=gr1h3] {$h^3$};

\end{scope}

\begin{scope}[xshift = 1cm, yshift=4cm]
\node[neuron, label={[xshift=-2.0cm, yshift=-0.6cm]{\large Pooling}},label={[xshift=-2.0cm, yshift=-1.2cm]{\large (avg/max)}}] (explosive1)  [above of=gr1h3] {$pool$};
\end{scope}

\end{scope}


\draw[orange,->] (gh1) -> node[orange,near start,left] {$w_l$} (gr1h2);
\draw[orange,->] (gh2) -> node[orange,near start,left] {$w_l$} (gr1h3);
\draw[orange,->] (gh3) -> node[orange,near start,left] {$w_l$} (gr1h4);

\draw[cyan,->] (gh2) -> node[cyan] {$w_m$} (gr1h2);
\draw[cyan,->] (gh3) -> node[cyan] {$w_m$} (gr1h3);
\draw[cyan,->] (gh4) -> node[cyan] {$w_m$} (gr1h4);

\draw[violet,->] (gh3) -> node[violet,near start,right] {$w_r$} (gr1h2);
\draw[violet,->] (gh4) -> node[violet,near start,right] {$w_r$} (gr1h3);
\draw[violet,->] (gh5) -> node[violet,near start,right] {$w_r$} (gr1h4);

\draw[->] (gr1h2) -> node[left] {} (explosive1);
\draw[->] (gr1h3) -> node[right] {} (explosive1);
\draw[->] (gr1h4) -> node[right] {} (explosive1);

\begin{scope}[xshift=11cm,yshift=0cm]

\begin{scope}[xshift=-1cm, yshift=-2cm]

\draw[step=2.0,black,thin,dashed] (0,0) grid (10,1.1);

\node at (1,0.8) {\large  $\scalar{v_1}\textcolor{red}{::} f(1)$};
\node at (3,0.8) {\large  $\scalar{v_2}\textcolor{red}{::} f(2)$};
\node at (5,0.8) {\large  $\scalar{v_3}\textcolor{red}{::} f(3)$};
\node at (7,0.8) {\large  $\scalar{v_4}\textcolor{red}{::} f(4)$};
\node at (9,0.8) {\large  $\scalar{v_5}\textcolor{red}{::} f(5)$};

\node[fill=white,inner sep=0pt,outer sep=0pt] at (2,0.25) {\large $next(1,2)$ };
\node[fill=white,inner sep=0pt,outer sep=0pt] at (4,0.25) {\large $next(2,3)$ };
\node[fill=white,inner sep=0pt,outer sep=0pt] at (6,0.25) {\large $next(3,4)$ };
\node[fill=white,inner sep=0pt,outer sep=0pt] at (8,0.25) {\large $next(4,5)$ };

\node[] (gh2) [xshift=5cm, yshift=-0.7cm] {\LARGE image $I$ as a set of weighted facts};

\end{scope}

\begin{scope}[xshift=0cm]

\node[gratom, right color=black!30!white] (gh1) {F$_{f(1)}$};
\node[gratom, right color=black!30!white] (gh2) [right of=gh1] {F$_{f(2)}$};
\node[gratom, right color=black!30!white] (gh3) [right of=gh2] {F$_{f(3)}$};
\node[gratom, right color=black!30!white] (gh4) [right of=gh3] {F$_{f(4)}$};
\node[gratom, right color=black!30!white, label={[xshift=1.5cm, yshift=-0.4cm]{\large Fact}},label={[xshift=1.5cm, yshift=-1cm]{\large nodes}}] (gh5) [right of=gh4] {F$_{f(5)}$};

\end{scope}

\begin{scope}[xshift=1cm, yshift=2cm]

\node[grbond] (gr1h2) [above of=gh2] {R$_{\alpha_1\theta_1}^{h}$};
\node[grbond] (gr1h3) [right of=gr1h2] {R$_{\alpha_1\theta_2}^{h}$};
\node[grbond,label={[xshift=1.5cm, yshift=-0.4cm]{\large Rule}},label={[xshift=1.5cm, yshift=-1cm]{\large nodes}}] (gr1h4) [right of=gr1h3] {R$_{\alpha_1\theta_3}^{h}$};

\end{scope}

\begin{scope}[xshift = 1cm, yshift=4cm]
\node[grexpl,label={[xshift=1.8cm, yshift=-0.6cm]{\large Aggregation}},label={[xshift=1.8cm, yshift=-1.1cm]{\large node}}] (explosive1)  [above of=gr1h3] {G$_{\alpha_1}^{h}$};
\end{scope}


\draw[orange,->] (gh1) -> node[orange,near start,left] {$w_l$} (gr1h2);
\draw[orange,->] (gh2) -> node[orange,near start,left] {$w_l$} (gr1h3);
\draw[orange,->] (gh3) -> node[orange,near start,left] {$w_l$} (gr1h4);

\draw[cyan,->] (gh2) -> node[cyan] {$w_m$} (gr1h2);
\draw[cyan,->] (gh3) -> node[cyan] {$w_m$} (gr1h3);
\draw[cyan,->] (gh4) -> node[cyan] {$w_m$} (gr1h4);

\draw[violet,->] (gh3) -> node[violet,near start,right] {$w_r$} (gr1h2);
\draw[violet,->] (gh4) -> node[violet,near start,right] {$w_r$} (gr1h3);
\draw[violet,->] (gh5) -> node[violet,near start,right] {$w_r$} (gr1h4);

\draw[gray,->] (gr1h2) -> node[left] {} (explosive1);
\draw[gray,->] (gr1h3) -> node[right] {} (explosive1);
\draw[gray,->] (gr1h4) -> node[right] {} (explosive1);

\end{scope}

\end{tikzpicture}

%% file: img/tikz/recurrent.tex
\begin{tikzpicture}
[transform shape,rotate=0, node distance=2.0cm and 2.0cm,
ar/.style={->,>=latex},
mynode/.style={
  draw, scale = 1.0,  minimum size=1cm, rounded corners,left color=white,
  minimum height=1cm,
  align=center
  }
]

\tikzstyle{neuron}  =  [rectangle, draw, scale = 1.0,  minimum size=1cm]
\tikzstyle{treenode}  =  [mynode, right color=black!30!white]
\tikzstyle{recnode}  =  [mynode, right color=red!30!white]
\tikzstyle{grbond}  =  [mynode, right color=black!30!white]
\tikzstyle{gratom}  =  [mynode]
\tikzstyle{grgroup} =  [mynode, right color=brown!30!white]
\tikzstyle{grexpl}  =  [mynode, right color=violet!30!white]
\tikzstyle{edgenode}  =  [thin, draw=black, align=center,fill=white,font=\small]

\node[treenode] (l1) {F$_{f(1)}$};
\node[treenode] (l2) [right of=l1] {F$_{f(2)}$};
\node[treenode] (l3) [right=2cm of l2] {F$_{f(k)}$};

\node[recnode] (h1) [below of=l1] {R$_{\alpha_1}^{h(1)}$};
\node[mynode,draw=black] (h0) [left of=h1] {A$_{h(0)}$};
\node[recnode] (h2) [below of=l2] {R$_{\alpha_1}^{h(2)}$};

\node[recnode] (h3) [below of=l3] {R$_{\alpha_1}^{h(k)}$};


\draw[orange,->] (l1) -> node[ left] {$\mathbf{W_f}$} (h1);
\draw[orange,->] (l2) -> node[ left] {$\mathbf{W_f}$} (h2);
\draw[orange,->] (l3) -> node[ left] {$\mathbf{W_f}$} (h3);

\draw[cyan,->] (h0) -> node[below] {$\mathbf{W_h}$} (h1);
\draw[cyan,->] (h1) -> node[below] {$\mathbf{W_h}$} (h2);
\draw[cyan,->, dashed] (h2) -> node[below] {$\mathbf{W_h}$} (h3);

\node[mynode, draw=white] (void) [right=0.5cm of h2] {$\dots$};
\node[mynode, draw=white] (void0) [right=0.5cm of l2] {$\dots$};

\end{tikzpicture}

%% file: img/tikz/recursive.tex
\begin{tikzpicture}
[transform shape,rotate=0, node distance=2.0cm and 2.0cm,
ar/.style={->,>=latex},
mynode/.style={
  draw, scale = 1.0,  minimum size=1cm, rounded corners,left color=white,
  minimum height=1cm,
  align=center
  }
]

\tikzstyle{neuron}  =  [rectangle, draw, scale = 1.0,  minimum size=1cm]
\tikzstyle{treenode}  =  [mynode, right color=red!30!white]
\tikzstyle{grbond}  =  [mynode, right color=black!30!white]
\tikzstyle{gratom}  =  [mynode]
\tikzstyle{grgroup} =  [mynode, right color=brown!30!white]
\tikzstyle{grexpl}  =  [mynode, right color=violet!30!white]
\tikzstyle{edgenode}  =  [thin, draw=black, align=center,fill=white,font=\small]

\node[grbond] (gh1) {F$_{n(1)}^{(0)}$};
\node[grbond] (gh2) [below=0.5cm of gh1] {F$_{n(2)}^{(0)}$};
\node[grbond] (gh3) [below=0.5cm of gh2] {F$_{n(3)}^{(0)}$};

\node[mynode, draw=white] (void) [above=0.2cm of gh1] {$\dots$};
\node[mynode, draw=white] (void) [below=0.2cm of gh3] {$\dots$};

\node[treenode] (l11) [right=3.5cm of gh1] {R$_{\alpha_1}^{{n(i)}^{(d-1)}}$};
\node[treenode] (l12) [right of=gh2] {R$_{\alpha_1}^{{n(j)}^{(1)}}$};
\node[treenode] (l13) [right=3.5cm of gh3] {R$_{\alpha_1}^{{n(k)}^{(d-1)}}$};

\node[treenode] (l22) [right=4cm of l12] {R$_{\alpha_1}^{{n(j)}^{(d)}}$};


\draw[orange,->] (gh1) -> node[ below] {$\mathbf{W_1}$} (l12);
\draw[cyan,->] (gh2) -> node[below] {$\mathbf{W_2}$} (l12);
\draw[violet,->] (gh3) -> node[below] {$\mathbf{W_3}$} (l12);

\draw[orange,dashed,->] (3,1) -> (l11);
\draw[cyan,dashed,->] (3,0) -> (l11);
\draw[violet,dashed,->] (3,-1) -> (l11);

\draw[orange,dashed,->] (3,-2) -> (l13);
\draw[cyan,dashed,->] (3,-3) -> (l13);
\draw[violet,dashed,->] (3,-4) -> (l13);

\draw[orange,->] (l11) -> node[ below] {$\mathbf{W_1}$} (l22);
\draw[cyan,dashed,->] (l12) -> node[below,near end] {$\mathbf{W_2}$} (l22);
\draw[violet,->] (l13) -> node[below] {$\mathbf{W_3}$} (l22);

\node[mynode, draw=white] (void) [right=1.5cm of l12] {$\dots$};

\end{tikzpicture}

%% file: img/tikz/GNNs.tex
\begin{tikzpicture}
[transform shape,rotate=0, node distance=2.0cm and 2.0cm,
ar/.style={->,>=latex},
mynode/.style={
  draw, scale = 1.0,  minimum size=1cm, rounded corners,left color=white,
  minimum height=1cm,
  align=center
  }
]

\tikzstyle{neuron}  =  [rectangle, draw, scale = 1.0,  minimum size=1cm]
\tikzstyle{treenode}  =  [mynode, right color=black!30!white]
\tikzstyle{recnode}  =  [mynode, right color=red!30!white]
\tikzstyle{recnode2}  =  [mynode, right color=magenta!20!white]
\tikzstyle{aggnode}  =  [mynode, right color=blue!30!white]
\tikzstyle{qnode}  =  [mynode, right color=blue!40!white]

\tikzstyle{grbond}  =  [mynode, right color=black!30!white]
\tikzstyle{gratom}  =  [mynode]
\tikzstyle{grgroup} =  [mynode, right color=brown!30!white]
\tikzstyle{grexpl}  =  [mynode, right color=violet!30!white]
\tikzstyle{edgenode}  =  [thin, draw=black, align=center,fill=white,font=\small]

\node[treenode] (g1) {F$_{n_1}$};
\node[treenode] (g2) [below left = 0.5 and 0.5cm of g1] {F$_{n_2}$};
\node[treenode] (g3) [below right = 1.5 and 0.1cm of g1] {F$_{n_3}$};
\node[treenode] (g4) [above left = 0.5 and 0.5cm of g1]{F$_{n_4}$};

\draw[black,-] (g1) -> (g2);
\draw[black,-] (g1) -> (g3);
\draw[black,-] (g2) -> (g3);
\draw[black,-] (g1) -> (g4);

\node[recnode] (h1) [right = 6.5cm of g1] {A$_{h(n_1)}^{(1)}$};
\node[recnode] (h2) [below left = 0.5 and 0.5cm of h1] {A$_{h(n_2)}^{(1)}$};
\node[recnode] (h3) [below right = 1.5 and 0.1cm of h1] {A$_{h(n_3)}^{(1)}$};
\node[recnode] (h4) [above left = 0.5 and 0.5cm of h1] {A$_{h(n_4)}^{(1)}$};

\node[recnode,scale=0.5] (r1) [below right=-1.2 and 4cm of g4] {R$^4_1$};

\node[recnode,scale=0.5] (r2) [below right=-1.7 and 3.5cm of g1] {R$^1_4$};
\node[recnode,scale=0.5] (r3) [below right=-1.1 and 3.5cm of g1] {R$^1_2$};
\node[recnode,scale=0.5] (r4) [below right=-0.5 and 3.5cm of g1] {R$^1_3$};

\node[recnode,scale=0.5] (r5) [below right=-1 and 4.2cm of g2] {R$^2_1$};
\node[recnode,scale=0.5] (r6) [below right=-0.4 and 4.2cm of g2] {R$^2_3$};

\node[recnode,scale=0.5] (r7) [below right=-0.3 and 4.2cm of g3] {R$^3_1$};
\node[recnode,scale=0.5] (r8) [below right=0.3 and   4.2cm of g3] {R$^3_2$};

\node[recnode,scale=0.5] (rh1) [below left = -1.2cm and 0.5cm of h1] {G1};
\node[recnode,scale=0.5] (rh2) [below left = -0.8cm and 0.5cm of h2] {G2};
\node[recnode,scale=0.5] (rh3) [below left = -0.2cm and 0.5cm of h3] {G3};
\node[recnode,scale=0.5] (rh4) [below left = -1.2cm and 0.5cm of h4] {G4};

\draw[gray,->] (r1) -> (rh4);
\draw[gray,->] (rh4) -> (h4);

\draw[gray,->] (r7) -> (rh3);
\draw[gray,->] (r8) -> (rh3);
\draw[gray,->] (rh3) -> (h3);

\draw[gray,->] (r2) -> (rh1);
\draw[gray,->] (r3) -> (rh1);
\draw[gray,->] (r4) -> (rh1);
\draw[gray,->] (rh1) -> (h1);

\draw[gray,->] (r5) -> (rh2);
\draw[gray,->] (r6) -> (rh2);
\draw[gray,->] (rh2) -> (h2);

\draw[black,loosely dotted] (h1) -> (h2);
\draw[black,loosely dotted] (h1) -> (h3);
\draw[black,loosely dotted] (h2) -> (h3);
\draw[black,loosely dotted] (h1) -> (h4);

\draw[violet,->] (g1) to[out=0,in=-150] node[ above, near end] {$W_2$} (h1);
\draw[violet,->] (g2) to[out=0,in=-150] node[ above] {$W_2$} (h2);
\draw[violet,->] (g3) -> node[ below] {$W_2$} (h3);
\draw[violet,->] (g4) -> node[ above, near start] {$W_2$} (h4);

\draw[orange,->] (g1) -> node[below] {$W_1$} (r1);

\draw[orange,->] (g4) -> node[below, near end] {$W_1$} (r2);
\draw[orange,->] (g2) -> node[ below] {$W_1$} (r3);
\draw[orange,->] (g3) to[out=70,in=190] node[below] {$W_1$} (r4);

\draw[orange,->] (g1) -> node[ below] {} (r5);
\draw[orange,->] (g3) -> node[ below] {$W_1$} (r6);

\draw[orange,->] (g1) to[out=-30,in=-170] node[above, near end] {$W_1$} (r7);
\draw[orange,->] (g2) to[out=-49,in=170] node[above] {$W_1$} (r8);

\node[recnode2] (hh1) [right = 4.5cm of h1] {A$_{h(n_1)}^{(n)}$};
\node[recnode2] (hh2) [below left = 0.5 and 0.5cm of hh1] {A$_{h(n_2)}^{(n)}$};
\node[recnode2] (hh3) [below right = 1.5 and 0.1cm of hh1] {A$_{h(n_3)}^{(n)}$};
\node[recnode2] (hh4) [above left = 0.5 and 0.5cm of hh1] {A$_{h(n_4)}^{(n)}$};

\draw[black,loosely dotted] (hh1) -> (hh2);
\draw[black,loosely dotted] (hh1) -> (hh3);
\draw[black,loosely dotted] (hh2) -> (hh3);
\draw[black,loosely dotted] (hh1) -> (hh4);

\draw[cyan,dashed,->] (h1) -> node[color=black, draw=white, fill=white] {\Large$\dots$} (hh1);
\draw[cyan,dashed,->] (h2) -> node[color=black, draw=white, fill=white] {\Large$\dots$} (hh2);
\draw[cyan,dashed,->] (h3) -> node[color=black, draw=white, fill=white] {\Large$\dots$} (hh3);
\draw[cyan,dashed,->] (h4) -> node[color=black, draw=white, fill=white] {\Large$\dots$} (hh4);

\draw[teal,dashed,->] (h1) -> node[color=black, draw=white, fill=white] {\Large$\dots$} (hh4);
\draw[teal,dashed,->] (h4) -> node[color=black, draw=white, fill=white] {\Large$\dots$} (hh1);
\draw[teal,dashed,->] (h2) -> node[color=black, draw=white, fill=white] {\Large$\dots$} (hh1);
\draw[teal,dashed,->] (h3) to[out=40,in=-170] node[color=black, draw=white, fill=white] {\Large$\dots$} (hh1);

\draw[teal,dashed,->] (h1) -> node[color=black, draw=white, fill=white] {\Large$\dots$} (hh2);
\draw[teal,dashed,->] (h3) -> node[color=black, draw=white, fill=white] {\Large$\dots$} (hh2);

\draw[teal,dashed,->] (h1) to[out=-30,in=-170] node[color=black, draw=white, fill=white] {}(hh3);
\draw[teal,dashed,->] (h2) -> node[color=black, draw=white, fill=white] {\Large$\dots$} (hh3);

\node[aggnode] (agg) [right = 2cm of hh1] {G$_{\alpha_n}^q$};
\draw[gray,->] (hh1) -> (agg);
\draw[gray,->] (hh2) -> (agg);
\draw[gray,->] (hh3) -> (agg);
\draw[gray,->] (hh4) -> (agg);

\node[qnode] (q) [right = 1.5cm of agg] {A$_q$};
\draw[blue,->] (agg) -> node[blue, below] {$W^{(n)}$} (q);

\end{tikzpicture}